\documentclass[11pt]{article}

\usepackage[preprint]{acl}
\usepackage{float}
\usepackage{booktabs}
\usepackage{times}
\usepackage{makecell}
\usepackage{latexsym}

\usepackage[T1]{fontenc}
\usepackage[utf8]{inputenc}

\usepackage{microtype}

\usepackage{inconsolata}

\usepackage{graphicx}
\usepackage{amsmath}
\title{Measuring Human Value Expression in Social Media Texts: Calibrated LLM Annotation and Encoder Transfer}

\author{
 \textbf{Maria Milkova\textsuperscript{1}},
 \textbf{Maksim Rudnev\textsuperscript{2}},
\\
\\
 \textsuperscript{1}Independent researcher, Lisbon, Portugal,
 \textsuperscript{2}University of Waterloo, ON, Canada
\\
 \small{
   \textbf{Correspondence:} {m.a.milkova@gmail.com}
 }
}

\begin{document}
\maketitle
\begin{abstract}
Measuring subjective constructs in naturally occurring social media text requires annotation procedures that are theoretically grounded, empirically validated, and transferable to an encoder model for scalable prediction. Using posts annotated according to Schwartz’s theory of basic human values, we investigate how different LLMs and prompting strategies, which we call annotation regimes, operationalize the expression of values in text. Beyond standard classification metrics, we evaluate structural alignment, annotation ambiguity, error patterns, and stability across repeated runs. We find that different LLMs produce different value interpretations, and iterative prompt calibration through error analysis reduces misattributions and improves alignment with expert annotations. Error patterns are further used to derive targeted expert-verification rules for corpus annotation. We transfer soft LLM labels to an encoder model for prediction, retaining information about ambiguity in value expression. Finally, a sensitivity analysis on more than one million posts shows that regime-specific annotation differences propagate into predicted levels of value expression, whereas standardized temporal dynamics and the direction of major event responses are more robust.
\end{abstract}

\section{Introduction}

Human value classification from text has emerged as an important task in computational social science and NLP \citep{kiesel-etal-2022-identifying, legkas2024hierocles} with social media providing an important source for studying value expression in communication and across online communities \citep{shifman2025expression, borenstein-etal-2025-investigating}. However, naturally occurring social media posts are often implicit, context-dependent, and highly heterogeneous, making value expression ambiguous. 

Existing work typically formulates value identification as a multi-label classification problem, where texts are annotated for the presence of predefined human values \citep{mirzakhmedova-etal-2024-touche23}. Importantly, the development of supervised classification models depends on the availability and quality of annotated data. While previous work uses value labels as gold standard annotations, recent advances show that value perception is partially in the eye of the beholder \citep{epstein2025measuring, falk-lapesa-2025-mining}. 

Large language models (LLMs) are increasingly used to annotate texts for human values \citep{starovolsky2025value, rodrigues2024beyond, de2025value}. However, relatively little attention has been paid to how different annotation configurations influence the interpretation of subjective constructs. Given the ambiguity of value expression and internal biases of LLMs, alternative models, prompt design and instruction languages may operationalize values differently and therefore produce different readings of the same texts. These challenges are further amplified in non-English settings, where linguistic and cultural contexts may diverge from those implicitly encoded in LLMs \citep{tao2024cultural, xu2024multilingualvalues}. 

When specific annotations are used to train classifiers, annotation biases can propagate into large-scale estimates of value prevalence and trends. This creates a central methodological challenge for subjective NLP tasks and computational social science. How should alternative LLM annotations be validated and transferred to downstream prediction when subjective constructs permit multiple plausible interpretations? And to what extent is large-scale prediction sensitive to the choice of annotations?

We study this problem in the context of human value annotation of naturally occurring non-English social media posts using Schwartz’s theory of basic human values \citep{schwartz1992universals}. We conceptualize combinations of models, prompts, and prompt languages as annotation regimes. Because aggregate classification metrics alone are insufficient for subjective annotation tasks, we also evaluate annotation regimes through structural alignment between values, confidence-ambiguity relations, and disagreement patterns. We further use recurrent disagreement patterns to derive targeted expert verification rules, enabling selective review of potentially problematic annotations. We then evaluate how calibrated LLM annotations can be transferred to an encoder model trained on soft labels. Finally, we provide a sensitivity analysis showing which forms of large-scale inference are most affected by annotation regime choice.
\section{Related work}

\subsection{Subjectivity in annotation }

Subjective NLP tasks often involve multiple plausible interpretations of the same text, resulting in disagreement among annotators \citep{Basile2021WeNT}. Recent work argues that this disagreement may reflect ambiguity, contextual variation, and differences in perspective \citep{plank-2022-problem, fleisig-etal-2024-perspectivist}. This view is consistent with the perspectivist paradigm, which regards annotation outcomes as situated judgments rather than direct observations of an objective ground truth \citep{davani-etal-2022-dealing}. Consequently, subjectivity should not be eliminated through aggregation, but explicitly identified and modeled as part of the annotation process \citep{homayounirad-etal-2025-will}. Recent studies of human value annotation similarly demonstrate that value attribution often depends on interpretation and may vary across annotators \citep{falk-lapesa-2025-mining, epstein2025measuring}.

\subsection{LLM annotation as measurement }

Interpretive variation does not disappear when reading and subsequent annotation is delegated to LLMs. As LLMs are increasingly used as text annotators, recent work argues that such coding systems should be treated as a measurement challenge, with particular attention to how concepts are operationalized and how annotations are validated \citep{tan-etal-2024-large, calderon-etal-2025-alternative, wallach2025evaluating, halterman2025codebook}. Recent evaluations suggest that LLMs do not consistently outperform human or supervised alternatives across annotation tasks, and their performance depends on the task, evaluation criterion, and annotation design \citep{kristensen2025chatbots, bojic2025comparing, calderon-etal-2025-alternative}.

Prior work shows that LLM annotation in subjective tasks is sensitive to prompt framing, role instructions, and perspective specification \citep{schaefer-etal-2025-demographics, vera2025llms, frohling-etal-2025-personas, radharapu-etal-2025-arbiters}. In multilingual settings, annotation outcomes may also vary with instruction language, standpoint, and cultural mismatch \citep{cui-etal-2025-bias}.

A related question is how alternative annotations should be compared and validated when the target construct is conceptually structured. In Schwartz’s theory, basic human values are organized into broader higher-order domains, which means that not all classification errors are substantively equivalent. In hierarchical and multi-label settings structured label spaces require specifically designed metrics \citep{kosmopoulos2015evaluation, amigo-delgado-2022-evaluating, plaud-etal-2024-revisiting, chen-etal-2023-unified}. 

Importantly, the introduction of LLMs into annotation workflows does not eliminate the need for human judgment \citep{pangakis2025keeping}. Recent work explores hybrid pipelines that combine model-generated labels with human verification, including selective review strategies based on uncertainty or confidence estimates \citep{wang2024human, kim-etal-2024-meganno, schroeder-etal-2025-human, gligoric-etal-2025-unconfident, rouzegar-makrehchi-2024-active}.

\section{Experimental Setup}

\subsection{Datasets }

Our empirical analysis is based on a corpus of Russian-language social media posts collected from VKontakte (VK), the largest Russian social media platform. The corpus, constructed by \citet{milkova-rudnev-2026}, includes both original posts and reposts and reflects naturally occurring users’ self-expression. After anonymization, filtering out spam, advertisements, and non-personal content, and selecting potentially value-expressive posts, the resulting corpus consists of 1,105,085 posts. This corpus serves as the sampling frame for three datasets used in our experiments and is subsequently used to evaluate the sensitivity of large-scale predictions to the annotation regime.

\begin{figure}[t]
\centering
\includegraphics[width=\linewidth]{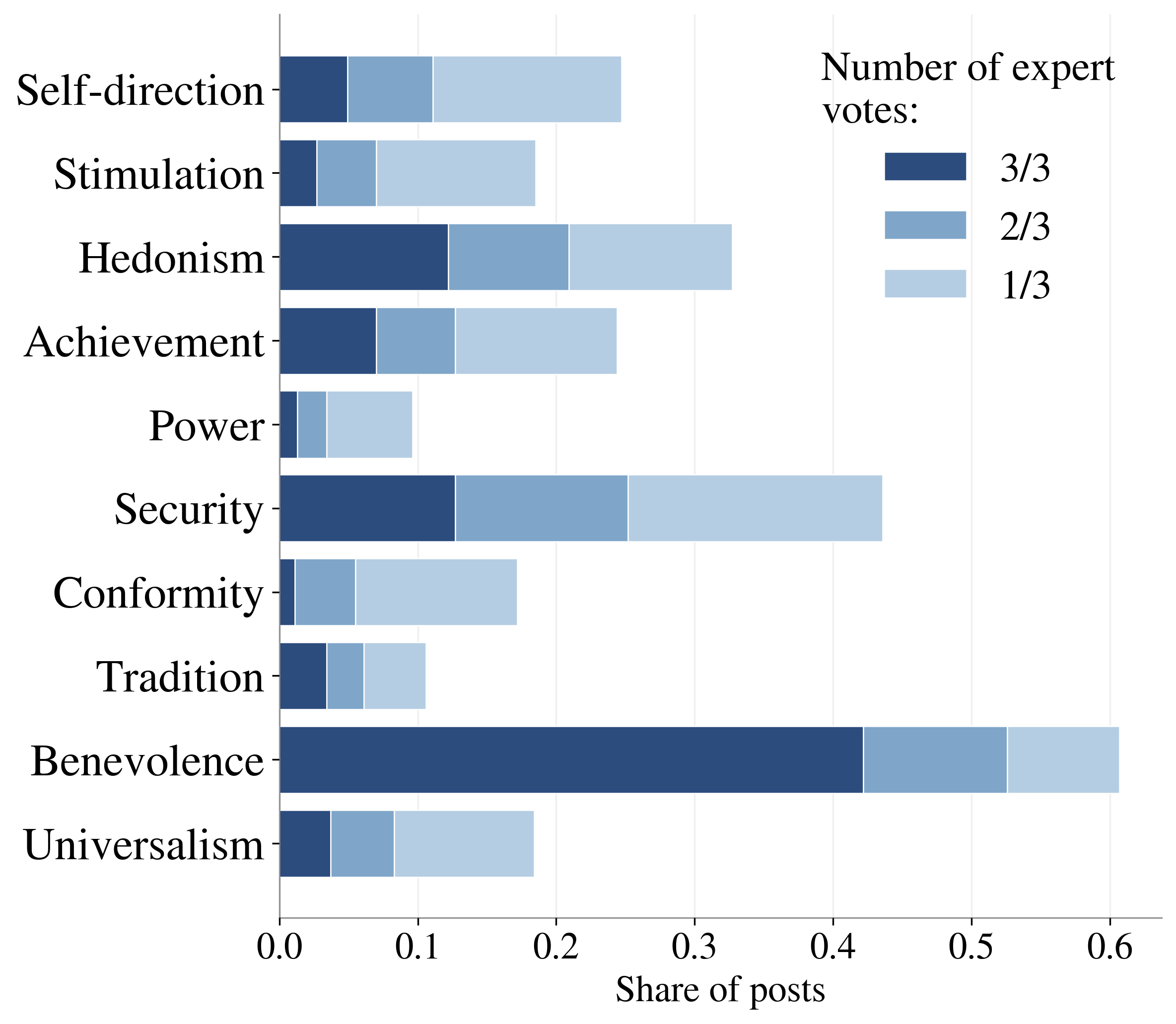}
\caption{Distribution of expert votes in Dataset-1 (N=1,000).}
\label{fig:expert_votes_dist}
\end{figure}

Dataset-1 consists of 1,000 posts annotated independently by three experts (with research backgrounds in human values) in a multi-label setting for the presence of each of the ten Schwartz values. Before the main annotation stage, experts completed a pilot annotation task consisting of 80 posts and wrote justifications for assigned values to confirm their conceptual alignment with value theory. Expert annotators (here and further for Dataset-2) were native Russian speakers, recruited through the authors’ academic networks and participated voluntarily without payment. Dataset-1  serves as a benchmark for evaluating LLM annotation regimes. Figure~\ref{fig:expert_votes_dist} illustrates the distribution of value labels and the corresponding levels of expert consensus (1/3, 2/3, and 3/3 votes). Overall expert agreement was moderate (Fleiss’ k = 0.6). The distribution is imbalanced across values, with Benevolence and Security being the most prevalent values, whereas Power and Conformity are the least represented. 

Dataset-2 consists of 2,000 posts annotated independently by a different group of three experts. For each of the ten values, 200 posts were randomly sampled and annotated in a binary setting, where experts judged only a single target value per post. The resulting positive-label rates and expert-vote distributions are shown in Appendix~\ref{app:expert_votes-dataset2}. We use this dataset to assess the robustness of the final LLM annotation regime. 

Dataset-3 consists of 24,000 posts used for large-scale LLM annotation and downstream encoder training.

\subsection{LLM annotation regimes }
We started with a calibration stage in which we evaluated three models: GPT-4, GPT-5.2, and Gemini-2.5-pro. These models were selected based on their strong performance in preliminary experiments on subjective annotation tasks. For each model, we tested different prompts (Table~\ref{tab:prompts_short}), which vary in the level of details in value definition, cultural framing, and contextual cues. Prompt variants were designed as successive stages of conceptual refinement. Full prompts are provided in the Appendix~\ref{app:prompts}.

\begin{table}[htbp]
\centering
\footnotesize
\begin{tabular}{p{2.1cm} p{4.2cm}}
\hline
\textbf{Prompt variant} & \textbf{Description} \\
\hline
Baseline & Short value definitions \\
Extended & More detailed conceptual definitions \\
Contextualized & Russian social media framing \\
Bias-calibrated & Definitions refined via error analysis \\
Bias-calibrated RU & Russian-language bias-calibrated prompt \\
\hline
\end{tabular}
\caption{Short description of prompt variants.}
\label{tab:prompts_short}
\end{table}

All LLM annotation experiments were conducted on Dataset-1. For each annotation regime (model-prompt configuration), every post was annotated five times using independently constructed random batches and a temperature of 0.1. Majority-vote aggregation was used for general evaluation. Repeated LLM annotations provided probabilistic labels (soft labels), allowing us to represent uncertainty in value assignment beyond majority-vote aggregation. The best-performing annotation regime was subsequently evaluated on Dataset-2.

\subsection{Evaluation framework}

We evaluated annotation outcomes along several complementary dimensions, such as general alignment, structural alignment, ambiguity analysis, and error structure.

\textit{General alignment} between expert and LLM annotations was evaluated using majority voting through precision-recall-F1 metrics, as well as through over-attribution ($FP / \text{ExpertNegatives}$) and under-attribution ($FN / \text{ExpertPositives}$) rates. However, these metrics evaluate values independently and therefore do not capture the conceptual structure of Schwartz’s theory. Because values are organized into four higher-order domains, different annotation disagreements are not substantively equivalent. For example, confusing Self-direction with Stimulation is conceptually less severe than confusing Self-direction with Security. 

\textit{Structural alignment} was evaluated using a domain-aware adaptation of Jaccard similarity. Exact value matches receive full similarity. Confusions between values belonging to the same higher-order domain receive partial similarity (0.5), reflecting their conceptual proximity within Schwartz’s value structure. Cross-domain mismatches receive zero similarity. The metric incorporates optimal matching between predicted and expert-assigned value sets to account for partial overlap in multi-label annotations. Full metric definition is provided in Appendix~\ref{app:jaccard}.

\textit{Ambiguity analysis} examined whether annotation confidence in repeated LLM annotations aligns with ambiguity in expert judgments. To evaluate this, we computed Spearman correlations between the number of positive expert votes (0–3) and the number of positive model annotations (0–5). 

Prompt calibration followed an iterative error-driven protocol. Over-attributed values were refined by tightening their definitions, whereas under-attributed values’ definitions were expanded to better capture implicit expressions. The calibrated prompt was supplemented with information about motivational relationships between values, derived from Schwartz’s theory. Refinement also included adding interpretive guidance, such as instructions to prioritize conceptual meaning over lexical cues, avoid assumptions beyond the text, account for subtle value expressions, and avoid over-attributing values in emotional texts. Calibration continued until further refinements failed to improve one evaluation dimension without degrading others. The best-performing calibrated regime was used in all subsequent analyses. The full bias-calibrated prompt is provided in Appendix ~\ref{app:final_prompt}.

We also assessed annotation stability by repeating the best-performing regime in two independent sets of five annotation runs and computing Cohen's $\kappa$ and the average label variation between runs.

\textit{Error profiling} was used to identify contexts in which the best-performing annotation regime was more likely to misattribute values. We analyzed value co-occurrence patterns for over- and under- attributed values together with annotation confidence. These analyses were then used to derive rules for targeted expert verification.

\subsection{Large-scale annotation and encoder transfer}

After establishing the best-performing annotation regime, we applied it to Dataset-3 for large-scale annotation. Cases flagged as potentially problematic through error-based rules were selectively verified by experts. Corrections were provided independently by a single expert per post to maximize review coverage under annotation-cost constraints. Experts used labels (0, 0.6, 1.0), where 0.6 denotes partial evidence of value expression.

The resulting annotated corpus was then used to train an encoder-based classifier (20,000/4,000 train/test split). We fine-tuned XLM-RoBERTa-large on soft labels. Full training details are provided in Appendix~\ref{app:xlm_training}. We evaluated predictions from two perspectives. First, we assessed predictive quality using Precision-Recall AUC on both the held-out test set and Dataset-1. Second, we examined whether the classifier reproduces the distribution of values by comparing cumulative distributions of expert and LLM annotations, and encoder-based predictions. We additionally assess whether encoder predictions preserve the graded ordering of value expression by computing Spearman correlations between encoder probabilities and the number of positive expert votes. We argue that evaluating continuous prediction scores is more appropriate than applying a fixed decision threshold because our goal is to preserve graded ambiguity in value expression.

\subsection{Large-scale prediction and sensitivity to annotation regime}

Fine-tuned XLM-RoBERTa large was then applied to the full corpus of 1,105,085 posts to obtain large-scale predictions of value expression. However, annotations of subjective constructs are inherently interpretive, and the rapidly evolving LLM landscape makes the choice of annotation regime partly contingent on the models available at a particular time. We therefore ask how strongly large-scale predictions and the resulting substantive conclusions depend on the annotation regime used to train the encoder. To evaluate this sensitivity, we additionally annotated Dataset-3 using the lowest-performing regime in our experiments and fine-tuned XLM-RoBERTa on this data.

We then compared the two sets of large-scale predictions and evaluated differences in predicted levels of value expression, as well as the robustness of temporal trends and event responses. To assess predicted-level sensitivity, we computed paired monthly differences between the two prediction sets, with confidence intervals estimated using a moving-block bootstrap. To assess whether regime differences affect temporal dynamics, we compared month-to-month changes in value-specific mean z-standardized prediction scores. We computed directional agreement in the signs of monthly changes and estimated value-specific regressions 
\[\Delta y^{\mathrm{LLM1}}_{v,t} = \alpha_v + \beta_v \Delta y^{\mathrm{LLM2}}_{v,t} + \epsilon_{v,t},\] using HAC confidence intervals for $\beta$. As a final sensitivity check, we examined responses to the Russian invasion of Ukraine on February 24, 2022, a major shock to Russian-language social media discourse. For each value and prediction set, we computed the event response as the difference between mean daily z-standardized scores in the 30-day post-event and pre-event windows.

\section{Results}
\subsection{Model-specific annotation biases}
\begin{figure}[t]
\centering
\includegraphics[width=0.8\linewidth]{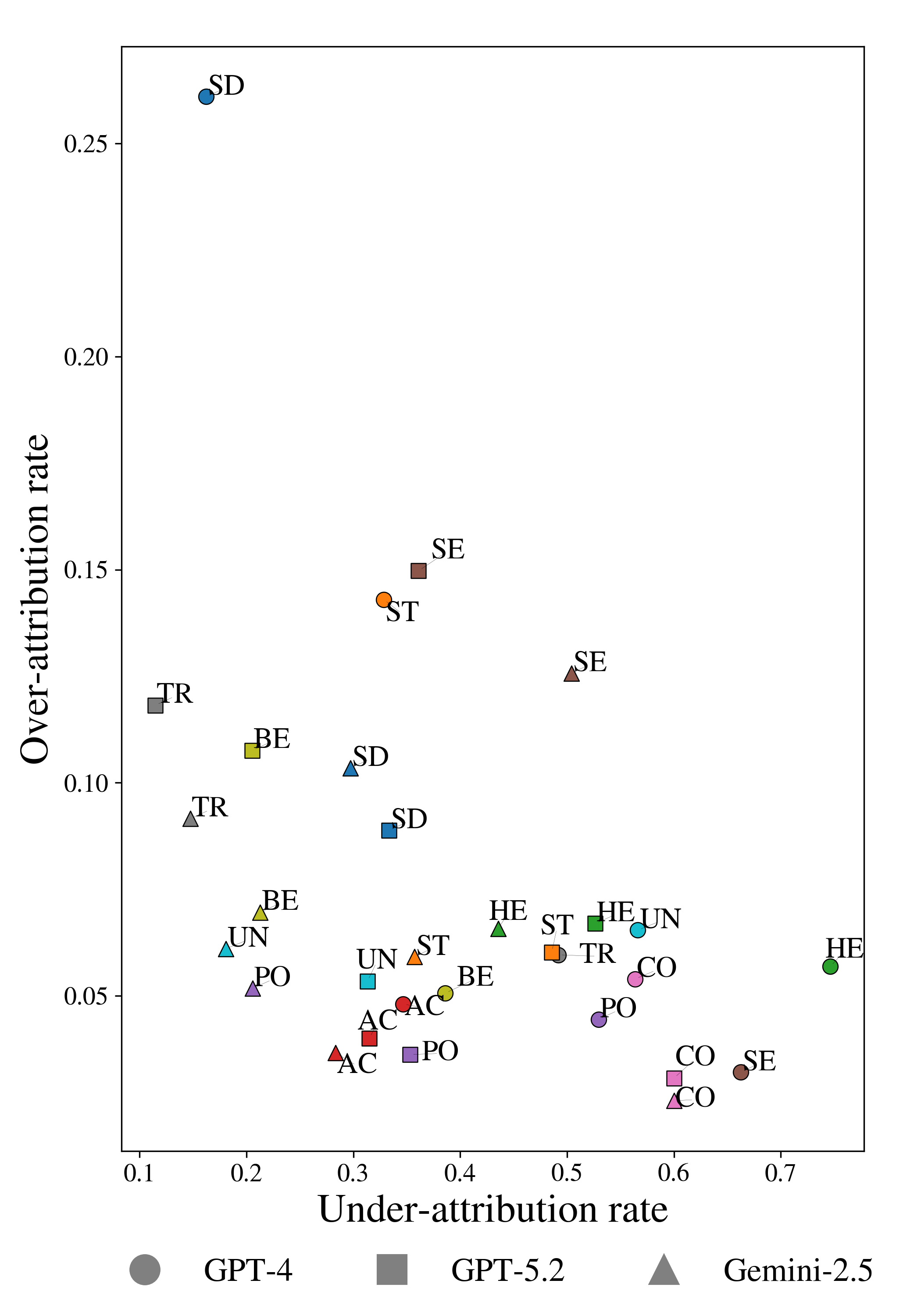}
\caption{Over- and under-attribution rates across values and models under the Baseline prompt.
\footnotesize
SD: Self-direction; ST: Stimulation; HE: Hedonism; AC: Achievement; PO: Power; SE: Security; CO: Conformity; TR: Tradition; BE: Benevolence; UN: Universalism.}
\label{fig:FP_FN_baseline}
\end{figure}

Under the Baseline prompt, newer models (GPT-5.2 and Gemini-2.5) achieve stronger alignment with expert annotations than GPT-4, with overall F1 increasing from 0.52 for GPT-4 to 0.65 and 0.66 for GPT-5.2 and Gemini, respectively (see Appendix~\ref{app:annotation_regimes} for detailed metrics).
However, different LLMs exhibit different value attribution biases under identical prompting conditions. As shown in Figure~\ref{fig:FP_FN_baseline}, under the Baseline prompt, GPT-4 over-attributes values belonging to the Openness to Change domain (Self-direction and Stimulation), whereas GPT-5.2 and Gemini over-attribute Conservation-related values, such as Security and Tradition. All models under-attribute Hedonism and Conformity.

\subsection{General performance across annotation regimes}

Prompt calibration alters annotation behavior. Across models, extending value definitions generally increases precision and reduces over-attribution, often at the cost of lower recall. However, models respond differently to refinement (see Figure~\ref{app:P-R-F1-across_model-prompts} in Appendix). Relative to the Baseline prompt, GPT-5.2 becomes more conservative after calibration, achieving higher precision (0.63→0.76) but missing more implicit value expressions (recall 0.67→0.58). In contrast, Gemini maintains a more balanced precision–recall trade-off (precision 0.64→0.72, recall 0.68→0.67). In Table~\ref{tab:prompt_calibration} we report overall F1 across annotation regimes. Detailed precision and recall metrics are provided in Appendix~\ref{app:annotation_regimes}) .

For example, a post expressing willingness to help and support a close person (\textit{“I promise I'll be there for you -— just let me know if you need me”}) was annotated by experts as Benevolence. Under the Baseline prompt, GPT-4 additionally attributed Self-direction, while GPT-5.2 and Gemini attributed Security. After calibration, all models converged on Benevolence only, eliminating these spurious co-attributions.

Russian-language prompting generally increases recall and over-attribution rates while reducing precision. Consequently, improvements in F1 observed for the Russian-language variants are driven by higher recall, which outweighs the accompanying loss in precision. Although the Russian-language GPT-5.2 regime achieves higher precision than Gemini (0.74 vs. 0.69), it also exhibits lower recall (0.65 vs. 0.72) (Appendix~\ref{app:annotation_regimes}).

\begin{table}[t]

\centering
\footnotesize
\setlength{\tabcolsep}{8pt}

\begin{tabular}{llccc}
\hline
\textbf{Model} & \textbf{Regime} & \textbf{F1} \\
\hline

GPT-4
& Baseline    & 0.52 \\
& Bias-calibrated     & 0.56 (+0.04) \\
& Bias-calibrated RU  & {0.58 (+0.06)} \\
\hline

GPT-5
& Baseline    & 0.65 \\
& Bias-calibrated     & 0.66 (+0.01) \\
& Bias-calibrated RU  & {0.69 (+0.04)} \\
\hline

Gemini
& Baseline    & 0.66 \\
& Bias-calibrated     & 0.70 (+0.04) \\
& Bias-calibrated RU  & {0.70 (+0.04)} \\
\hline

\end{tabular}

\caption{Effect of prompt calibration. Values in parentheses indicate changes relative to the baseline configuration of the corresponding model.}
\label{tab:prompt_calibration}
\end{table}

\subsection{Structural alignment}

\begin{figure}[htbp]
\centering
\includegraphics[width=\linewidth]{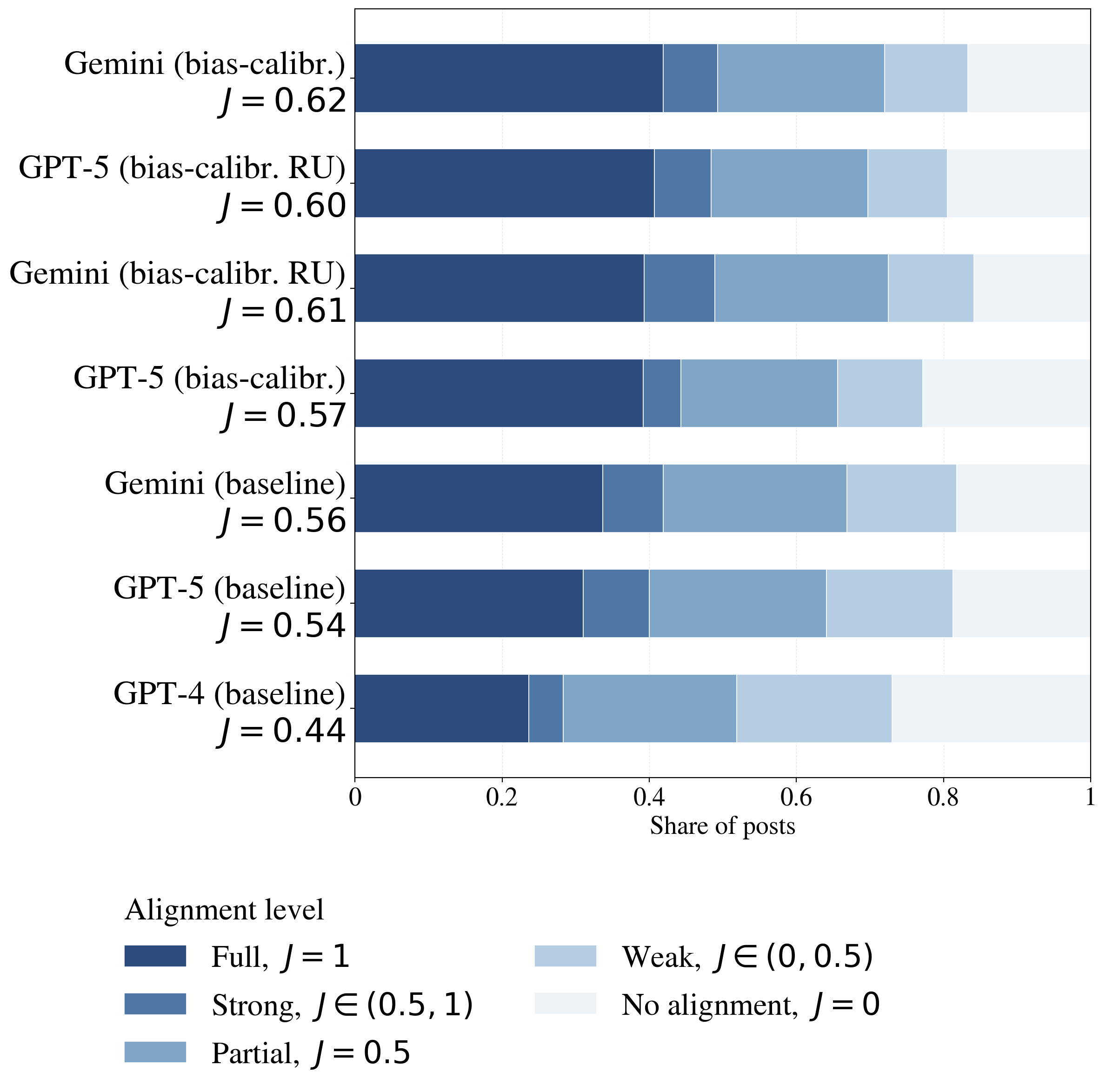}
\caption{Domain-aware Jaccard similarity across several annotation regimes.}
\label{fig:dp_jaccard_distribution}
\end{figure}

We then evaluate annotation regimes using domain-penalized Jaccard similarity to compare their structural alignment. To better characterize alignment structure, we decompose Jaccard scores into five categories: full alignment (J = 1), strong alignment (0.5 < J < 1), partial alignment (J = 0.5), weak alignment (0 < J < 0.5), and no alignment (J = 0). Figure~\ref{fig:dp_jaccard_distribution} shows that prompt refinement improves structural alignment, with Gemini under the bias-calibrated prompt achieving the highest structural alignment with expert annotations (mean Jaccard increasing from 0.56 to 0.62). Jaccard scores for all annotation regimes provided in Appendix~\ref{app:jaccard_metrics}.

\subsection{Ambiguity in value annotation}

Across annotation regimes, we observe significant positive correlations between expert and model vote counts. Correlations increased across prompt refinements from 0.47 to 0.54 for GPT-4, from 0.61 to 0.64 for GPT-5.2, and from 0.64 to 0.67 for Gemini (see Appendix~\ref{app:correlations_all_regimes}). Across values, correlations ranged from 0.40 for Conformity to 0.83 for Benevolence (for Gemini bias-calibrated prompt, see Appendix~\ref{app:final_regime_ro}). These results suggest that, if the number of expert votes is treated as a proxy for ambiguity in value expression, repeated LLM annotations capture this variation.

\subsection{Final calibrated regime}
Based on the combined evaluation across aggregate metrics, over-attribution profiles, and structural alignment, we select Gemini with the English bias-calibrated prompt as the final annotation regime. Table~\ref{tab:final_regime} shows the final performance metrics of the selected regime. Performance varies across values with Benevolence exhibiting the strongest alignment with expert annotations (F1=0.88), followed by Achievement (0.71), Tradition (0.67), and Universalism (0.67). In contrast, Conformity value remains difficult to identify (0.39). 
\begin{table}[htbp]
\centering
\footnotesize
\setlength{\tabcolsep}{8pt}

\begin{tabular}{lccccc}
\hline
\textbf{Value} & \textbf{P} & \textbf{R} & \textbf{F1} & \textbf{OA} & \textbf{UA} \\
\hline

Self-direction & 0.53 & 0.70 & 0.61 & 0.08 & 0.30 \\
Stimulation    & 0.48 & 0.54 & 0.51 & 0.04 & 0.46 \\
Hedonism       & 0.77 & 0.49 & 0.60 & 0.04 & 0.51 \\
Achievement    & 0.72 & 0.70 & 0.71 & 0.04 & 0.30 \\
Power          & 0.43 & 0.71 & 0.53 & 0.03 & 0.29 \\
Security       & 0.71 & 0.50 & 0.59 & 0.07 & 0.50 \\
Conformity     & 0.57 & 0.29 & 0.39 & 0.01 & 0.71 \\
Tradition      & 0.55 & 0.85 & 0.67 & 0.05 & 0.15 \\
Benevolence    & 0.93 & 0.83 & 0.88 & 0.07 & 0.17 \\
Universalism   & 0.57 & 0.81 & 0.67 & 0.06 & 0.19 \\

\hline
\end{tabular}

\caption{Performance of Gemini bias-calibrated annotation regime. OA = over-attribution rate, UA = under-attribution rate.}
\label{tab:final_regime}
\end{table}

Table~\ref{tab:final_regime} also reports asymmetric error rates. Over-attribution rates (OA) range from 0.01 (for Conformity) to 0.07-0.08 (for Security, Benevolence, and Self-direction). The highest under-attribution rates (UA) are observed for Conformity (0.71) and Hedonism (0.51). These asymmetric error profiles motivate the targeted verification rules discussed in Section 4.6.

To assess robustness, we evaluated the selected regime on Dataset-2. Overall F1 decreased from 0.70 to 0.64 under different annotator group and expert annotation protocol. Importantly, the difficulty of value identification remains largely consistent across datasets, with Benevolence, Tradition, and Achievement exhibiting the strongest alignment in both settings, and Conformity the weakest. Precision, Recall, F1, over- and under- attribution rates for Dataset-2 are reported in Appendix~\ref{app:dataset2}.

We additionally evaluate the stability of repeated annotations produced by the final regime. Across independently constructed annotation batches (5×5 repeated runs), the regime demonstrates high inter-run consistency (Cohen’s k = 0.88) with low average variation in vote counts (mean delta votes = 0.19), indicating that repeated annotations remain stable. 

\subsection{Error structure and rules for expert verification}
\begin{figure}[H]
\centering
\includegraphics[width=\linewidth]{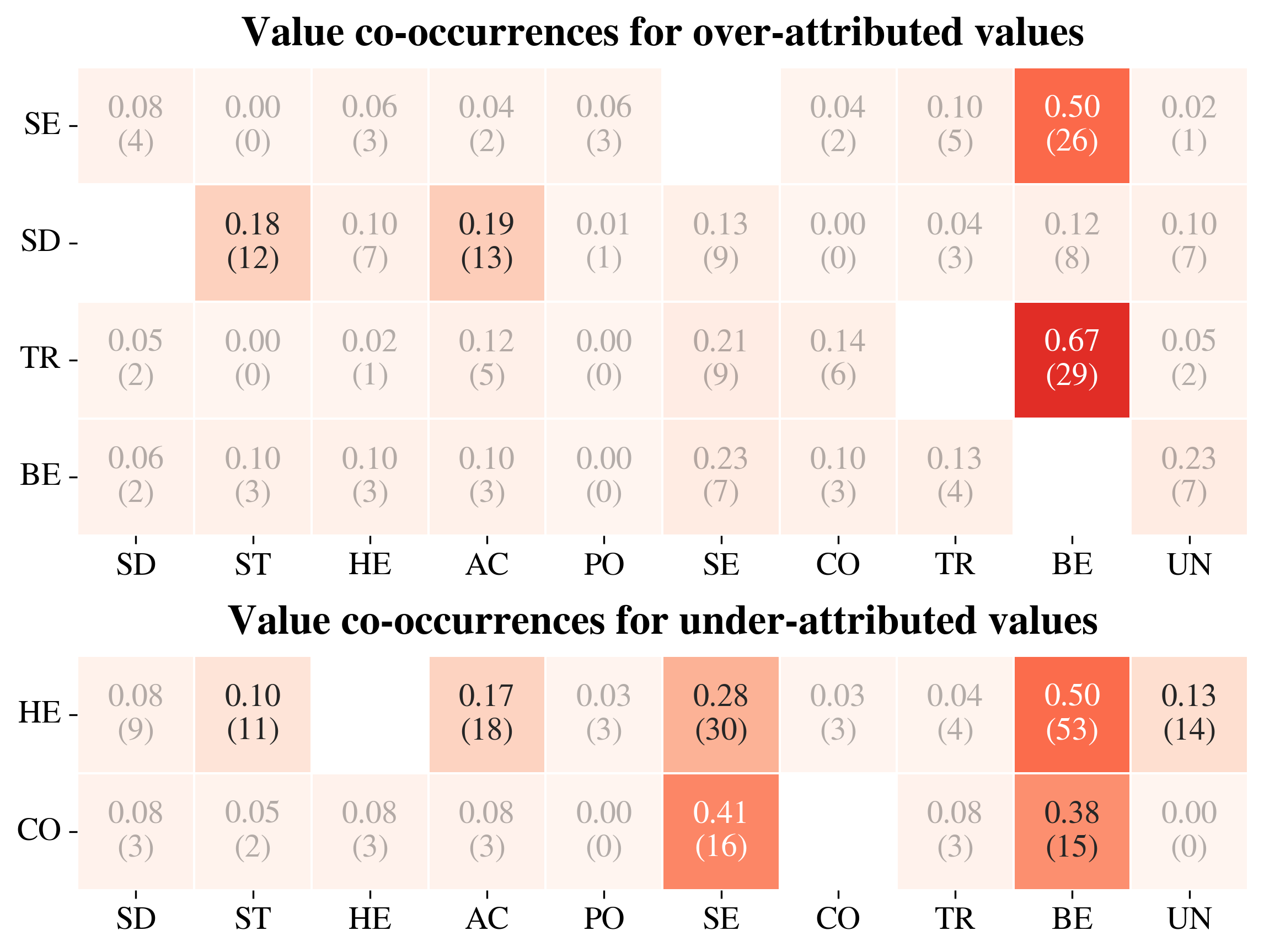}
\caption{Contexts associated with over- and under- attributed values.}
\label{fig:value_co-occurrences_dataset1}
\end{figure}

As shown in Table~\ref{tab:final_regime} and the robustness check on Dataset-2 (Appendix~\ref{app:dataset2}), several values continue to exhibit higher over- and under-attribution rates. To improve the reliability of large-scale annotation, we introduced targeted expert verification. To form the rules for verification, we first identified values associated with relatively high over-attribution rates (Security, Benevolence, Self-direction, and Tradition) and under-attribution rates (Hedonism and Conformity). To understand contexts in which these errors occurred most frequently, we then examined value co-occurrence patterns associated with over- and under-attributed cases (Figure~\ref{fig:value_co-occurrences_dataset1}). 

Figure~\ref{fig:value_co-occurrences_dataset1} shows that, for example, over-attribution of Security was frequently associated with co-occurring Benevolence, whereas under-attribution of Hedonism and Conformity was concentrated in posts co-occurring with Benevolence and Security. We additionally applied the same co-occurrence analysis to Dataset-2 and used the resulting Dataset-2-specific patterns to complement the final set of verification contexts (Appendix~\ref{app:dataset2_fp_fn}).

We further observed that misattributions concentrate disproportionately in lower-confidence annotations. Over- attributed cases are substantially more likely for unstable positive predictions (3–4 out of 5 votes) than for unanimous predictions (OR = 4.5, p < 0.001), while under- attributed cases are more likely for low-confidence negative predictions (1–2 votes vs. 0 votes; OR = 8.4, p < 0.001).

We used both co-occurrence patterns and annotation confidence to derive targeted verification rules. We excluded pairs of conceptually adjacent values and retained only those co-occurrences that reflect broader or cross-domain confusions, because confusions between conceptually adjacent values are less consequential from the perspective of Schwartz’s value structure. The full list of target–context pairs and verification rules are provided in Appendix~\ref{app:expert_verification_rules}.
\subsection{Transferring to encoder model}
We then applied the best-performing annotation regime to Dataset-3 for large-scale annotation. The verification rules identified 5,225 potentially problematic annotations. Due to annotation-cost constraints, we randomly sampled 2,000 cases for expert verification proportionally to the size of each flagged subgroup. Expert review resulted in 962 corrections of target-value annotations (on average, 64\% for over-attribution rules and 24\% for under-attribution rules). Detailed statistics for flagged, sampled, and corrected cases by value group are provided in Appendix~\ref{app:expert_verification}.

The resulting dataset is then used to fine-tune XLM-RoBERTa-large on soft labels. The distribution of soft labels in the training data is reported in Appendix~\ref{app:train_data_distribution}. 

Table~\ref{tab:prauc_comparison} reports Precision-Recall AUC for predictions. For most values, the performance of the encoder model relative to expert annotations is comparable to or higher than the performance observed for Gemini relative to experts. Macro PR-AUC equals 0.55 for Gemini–Experts, 0.60 for XLM-RoBERTa–Experts, and 0.69 for XLM-RoBERTa–Gemini.
\begin{table}[t]
\centering
\small
\setlength{\tabcolsep}{5pt}
\begin{tabular}{lccc}
\hline
\textbf{Value} &
\makecell{\textbf{Gemini-}\\Experts} &
\makecell{\textbf{XLM-}\\Experts} &
\makecell{\textbf{XLM-}\\Gemini} \\
\hline

Self-direction & 0.54 & 0.59 & 0.74 \\
Stimulation & 0.41 & 0.44 & 0.65 \\
Hedonism & 0.59 & 0.68 & 0.67 \\
Achievement & 0.69 & 0.73 & 0.81 \\
Power & 0.39 & 0.47 & 0.52 \\
Security & 0.59 & 0.72 & 0.72 \\
Conformity & 0.26 & 0.23 & 0.36 \\
Tradition & 0.51 & 0.67 & 0.80 \\
Benevolence & 0.92 & 0.94 & 0.94 \\
Universalism & 0.61 & 0.56 & 0.71 \\
\hline
\textbf{Macro Average} & \textbf{0.55} & \textbf{0.60} & \textbf{0.69} \\
\hline
\end{tabular}
\caption{Precision--Recall AUC across experts, Gemini and XLM--RoBERTa predictions.}
\label{tab:prauc_comparison}
\end{table}

\begin{figure}[t]
\centering
\includegraphics[width=0.8\linewidth]{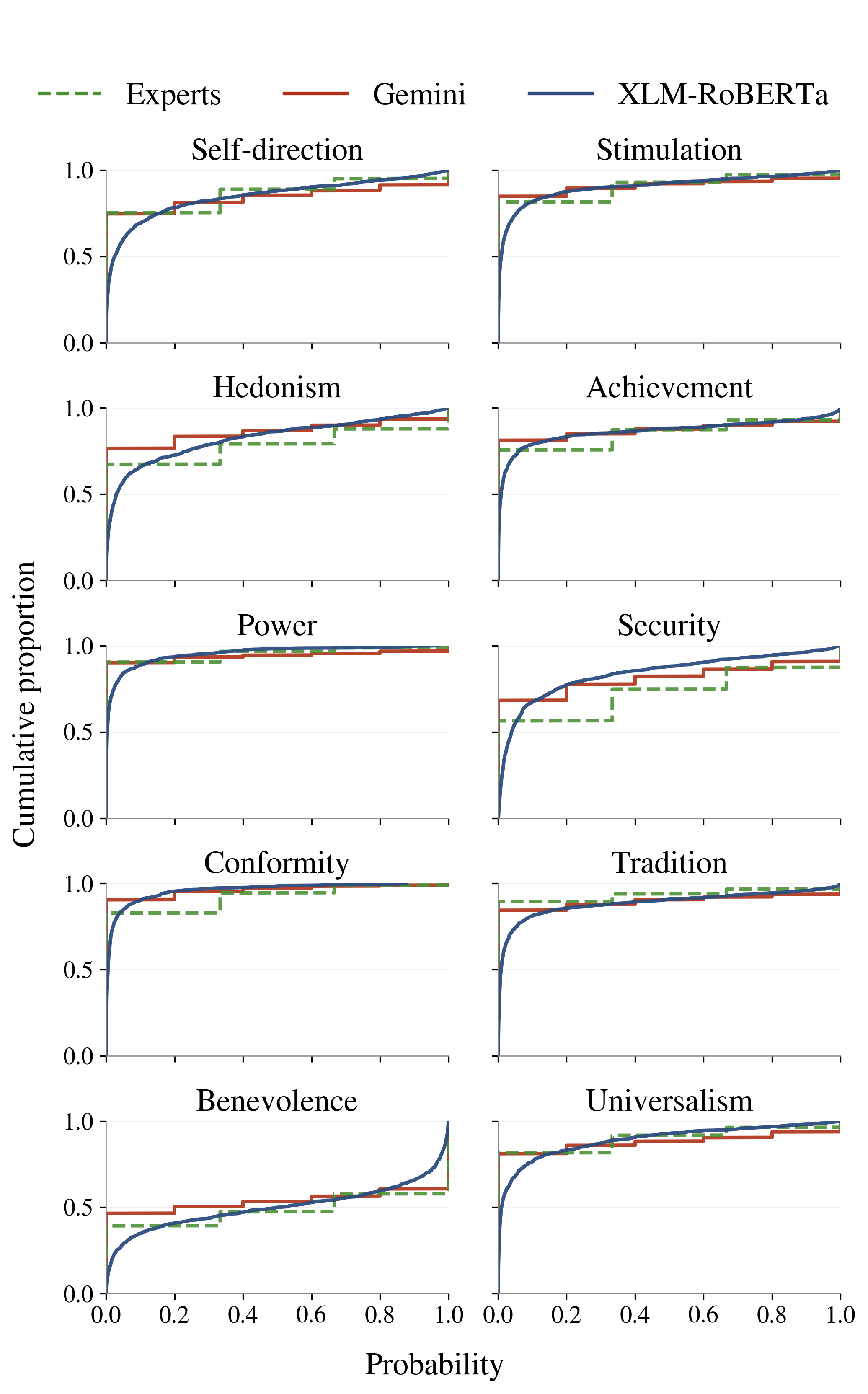}
\caption{Cumulative distribution functions.}
\label{fig:bert-llm-experts-bce}
\end{figure}

Figure~\ref{fig:bert-llm-experts-bce} compares cumulative prediction distributions across experts, Gemini, and XLM-RoBERTa. The encoder model largely reproduces the overall distributional structure of expert and LLM annotations, but smooths discrete high-confidence positive annotations.
To assess whether encoder predictions preserve graded value expression, we computed Spearman correlations between encoder probabilities and the number of positive expert votes. For all values, correlations are positive and significant (p < 0.001), ranging from 0.33 for Conformity to 0.81 for Benevolence (Table~\ref{tab:spearman_values}).
\begin{table}[t]
\centering
\footnotesize
\setlength{\tabcolsep}{6pt}
\begin{tabular}{lc}
\hline
\textbf{Value} & \textbf{Spearman $\rho$} \\
\hline

Self-direction & 0.545 \\
Stimulation & 0.444 \\
Hedonism & 0.624 \\
Achievement & 0.600 \\
Power & 0.368 \\
Security & 0.597 \\
Conformity & 0.333 \\
Tradition & 0.443 \\
Benevolence & 0.813 \\
Universalism & 0.532 \\
\hline
\end{tabular}
\caption{Spearman correlations between XLM--RoBERTa probabilities and number of positive expert votes.}
\label{tab:spearman_values}
\end{table}

These results suggest that the encoder model successfully transfers the interpretation of value categories and preserves part of the ambiguity present in expert judgments. Performance on the held-out test dataset is reported in Appendix~\ref{app:xlm_results}.

\subsection{Downstream robustness of social inference}
To analyze the sensitivity of predictions to underlying annotations, we compared large-scale predicted scores produced by encoders fine-tuned on GPT-4 annotations (baseline prompt) and Gemini-2.5 annotations (bias-calibrated prompt). These two annotation regimes were selected as the most different from each other, with the largest difference in aggregate performance (overall F1 of 0.52 vs. 0.70) and different value attribution patterns (see Appendix~\ref{app:train_annotation_differences}).

The two predictions have different monthly aggregate levels of value expression. The Gemini-based encoder assigned lower scores to Self-direction (-40.5\%) and Stimulation (-40.2\%) relative to the GPT-based encoder, with Gemini never exceeding GPT in any month for these values. In contrast, the Gemini-based encoder assigned consistently higher scores to Security (+64.9\%), Tradition (+41.5\%), and Benevolence (+34.7\%), with Gemini exceeding GPT in all months for these values (See Appendix~\ref{app:sensitivity}). These shifts correspond to regime-specific value attribution patterns observed during annotation.

\begin{figure}[htbp]
\centering
\includegraphics[width=0.7\linewidth]{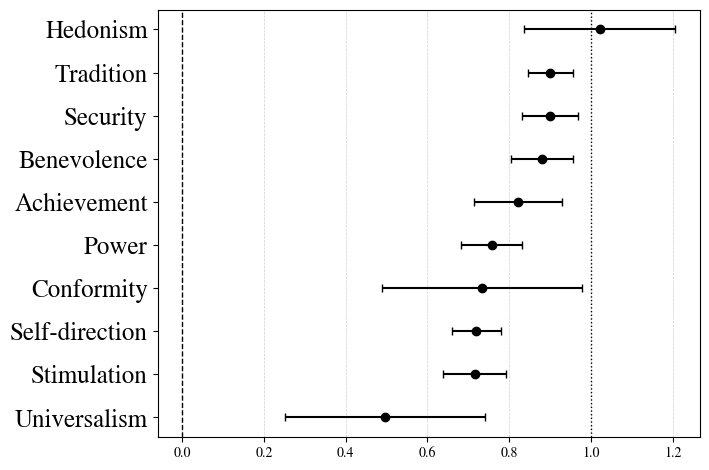}
\caption{$\beta$ coefficient from per-value monthly regressions of $\Delta z_{\mathrm{Gemini}}$ on $\Delta z_{\mathrm{GPT}}$, with 95\% HAC CIs.}
\label{fig:trend_change_scale_agreement_by_value}
\end{figure}

Temporal dynamics were more robust. In per-value monthly regressions of $\Delta z_{\mathrm{Gemini}}$ on $\Delta z_{\mathrm{GPT}}$, all $\beta$ coefficients were positive, indicating that increases and decreases in the GPT-based series corresponded to increases and decreases in the Gemini-based series. Gemini-based predictions usually attenuated month-to-month fluctuations relative to the GPT-based regime. Alignment was strongest for Security and Tradition ($\beta \approx .90$) and weakest for Universalism ($\beta = .50$) and Conformity ($\beta = .73$), see Figure~\ref{fig:trend_change_scale_agreement_by_value} and Appendix~\ref{app:sensitivity} for more details. Across values, 70.7-88.5\% of monthly changes were aligned in direction, the median value-level agreement was 85.6\%.

\begin{figure}[htbp]
\centering
\includegraphics[width=0.7\linewidth]{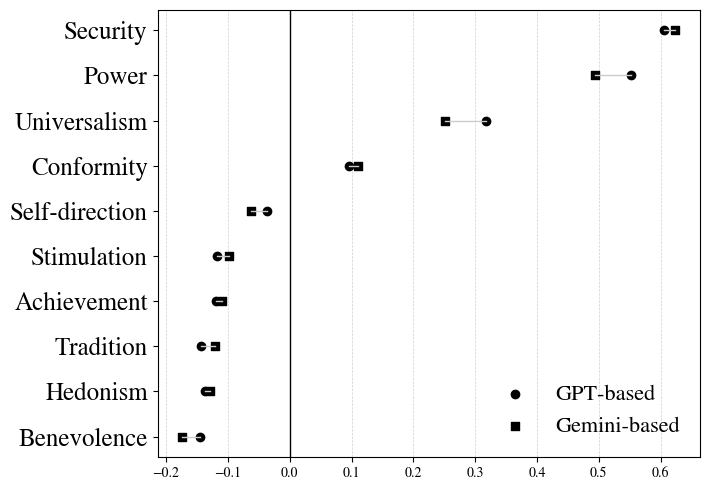}
\caption{Event response: post-event mean $-$ pre-event mean using 30-day windows.}
\label{fig:event_response_event_response_by_value}
\end{figure}

Analyzing event-response sensitivity, we evaluate differences in reaction to Russia’s Invasion of Ukraine. The response was highly consistent across regimes, with all ten values changed in the same direction. Both regimes identified the strongest post-event increases for Security, Power, Universalism, and Conformity. Benevolence showed the largest decrease. The largest regime differences in response magnitude were observed for Universalism and Power (See Figure~\ref{fig:event_response_event_response_by_value}).

\subsection{Discussion}

Our findings contribute to the growing literature on human value detection by shifting attention from classification performance alone to the problem of value measurement. Previous research has shown that values can be detected in text and that LLMs can be used as annotators. However, when values are inferred from naturally occurring everyday self-expression, the problem of value operationalization becomes central, and the choice of annotation regime becomes a measurement decision.

Our results show that different LLMs produce different value interpretations. Prompt calibration can reduce some spurious attributions and make the annotation regime more theoretically constrained and more diagnostically transparent.

Importantly, repeated LLM annotations provide useful information about uncertainty in value expression. This supports the use of soft labels for encoder-based models. In this setting, ranked probabilities produced by the encoder model retain information about the ambiguity in value expression.

Scalable ambiguity-aware prediction may be particularly useful for tracking aggregate patterns of value expression in social media discourse and examining how these patterns change over time and in response to major events.

Sensitivity results show that annotation biases propagate to absolute estimates of value levels. While annotation calibration should still be conducted carefully and guided by the observed error structure, final large-scale conclusions should take into account the subjectivity of annotations, both by LLMs and by experts. Thus, conclusions about value prevalence and absolute value scores should be evaluated with caution. Using z-standardized scores, the direction and broad pattern of temporal changes and event responses were more robust to annotation regime, although their magnitude remained sensitive to the underlying annotations. Ipsatized scores should also be used with caution, taking into account possible value-specific annotation biases. Importantly, the values that remain most sensitive to regime are Conformity, Hedonism, Universalism, and Power.

\section*{Limitations}

Our study is limited to Russian-language data. Although the proposed calibration and validation framework is intended to be language-independent, annotation behavior and value interpretation may differ across linguistic and cultural settings. Future work should therefore evaluate the proposed framework across additional languages and cultural contexts. Also, our analysis is based exclusively on textual content. In social media posts, value expression may depend on visual context, memes, or attached media that are not recoverable from text alone. More complete identification of value expression will require multimodal modeling.

Another limitation is that we base our study on ten basic human values and do not utilize the refined theory of nineteen values \citep{schwartz2012refining}. As a result, the proposed annotation scheme does not distinguish between more specific value types, restricting a more nuanced understanding of value expression.

Our annotations reflect theory-informed expert interpretation. They should not be interpreted as universal judgments of how ordinary users perceive value expression. 

The relatively small size of Dataset-1 may limit the completeness of prompt calibration and error profiling for rare values. Dataset-2 provides an additional robustness check with a different expert group, but it uses a single-value binary protocol, which may not fully capture interactions between simultaneously expressed values. This design was motivated by annotation cost, since evaluating all ten values simultaneously requires substantial expertise in value theory and is difficult to scale through crowd-based annotation platforms.

Importantly, our initial motivation was to develop and validate an annotation regime capable of capturing value expression. For this reason, we did not evaluate the sensitivity of annotation outcomes to each individual prompt addition. 

We also did not evaluate few-shot prompting for LLM annotation. Few-shot examples may improve alignment with expert annotations, but selecting representative posts for highly heterogeneous value expressions is itself non-trivial and may introduce additional contextual biases tied to selected examples.

During the expert verification stage, only 40\% of flagged cases were reviewed, and experts evaluated only the target value rather than the full multi-label annotation profile. In addition, each reviewed post was assessed by a single expert. Although this design reflects annotation constraints for large-scale subjective tasks, additional expert review and multi-expert verification may further improve annotation quality.

\section*{Ethics Statement}

This study analyzes social media posts collected from open user profiles only. User and post identifiers were hashed prior to analysis. Due to the potential sensitivity and re-identification risk of social media content, we do not release the source dataset. Model predictions reflect value expression in individual texts and should not be interpreted as direct measurements of users’ personal values.

\bibliography{references}

\clearpage
\onecolumn
\appendix
\label{sec:appendix}
\section{Prompt Specifications}
\label{app:prompts}

\subparagraph{Baseline prompt}
\small

\begin{quote}
You are a professional annotator tasked with categorizing social media posts.
Assign a binary code to each post - whether (code 1) or not (code 0) the post reflects the Values listed below:

Self-direction: Independent thought and action — choosing, creating, and exploring.

Stimulation: Excitement, novelty, and challenge in life.

Hedonism: Pleasure and sensuous gratification for oneself.

Achievement: Personal success through demonstrating competence according to social standards.

Power: Social status and prestige, wealth, control or dominance over people and resources.

Security: Safety, harmony, strong government, and stability of society, relationships, and self.

Conformity: The restraint of actions, inclinations, and impulses that are likely to upset or harm others and violate social expectations or norms.

Tradition: Respect, commitment, and acceptance of the customs and ideas that traditional culture or religion provides.

Benevolence: Preservation and enhancement of the welfare of people with whom one is in frequent personal contact.

Universalism: Understanding, appreciation, tolerance, and protection for the welfare of all people and of nature.

\end{quote}
\small

\subparagraph{Extended prompt}
\small

\begin{quote}
You are a professional annotator of social media posts.
Assign a binary code to each post - whether (code 1) or not (code 0) the post reflects basic human values.
Values are an expression of the importance of an object, phenomenon, or quality. The Values are listed below:

Self-direction: importance of independent thought and action, choosing, creating, and exploring, curious, autonomous.

Stimulation: Excitement, novelty, and challenge in life, daring

Hedonism: Pleasure and sensuous gratification for oneself, enjoying life, self-indulgent

Achievement: Personal success through demonstrating competence according to social standards, ambitious, capable, influential, social recognition

Power: Social status and prestige, wealth, control or dominance over people and resources, authority, social power

Security: Safety, harmony, strong government, and stability of society, relationships, and self, social order, family security, national security, clean, healthy

Conformity: The restraint of actions, inclinations, and impulses that are likely to upset or harm others and violate social expectations or norms, obedient, self-discipline, politeness, responsible

Tradition: Respect, commitment, and acceptance of the customs and ideas that traditional culture or religion provides, humble, devout, moderate, subordination

Benevolence: Preservation and enhancement of the welfare of people with whom one is in frequent personal contact, helpful, honest, forgiving, responsible, loyal, true friendship, mature love

Universalism: Understanding, appreciation, tolerance, and protection for the welfare of all people and of nature, broadminded, social justice, equality, world at peace, protecting the environment.

When assigning values, prioritize conceptual meaning over surface-level lexical cues.
Don't overestimate emotional texts.
Do NOT make assumptions about the author and intentions beyond the text.

Power, achievement, hedonism, stimulation, and self-direction primarily focus on personal interests and characteristics.
Benevolence, universalism, tradition, conformity, and security are primarily concerned with how one relates socially to others and affects their interests.
Power, achievement, tradition, conformity, and security serve to cope with anxiety due to uncertainty in the social and physical world.
Hedonism, stimulation, self-direction, universalism, and benevolence express anxiety-free motivations. Security and conformity aim to avoid or overcome actual or potential danger.
Self-direction and close values motivate intrinsically rewarding social, intellectual, and emotional opportunities.

\end{quote}
\small

\subparagraph{Contextualized prompt}
\small
\begin{quote}
This prompt is identical to the Extended prompt, with the only difference being at the beginning: You are a professional annotator of Russian social media VKontakte born in Russia and living in Russia.
\end{quote}
\small
\subparagraph{Bias-calibrated prompt}
\label{app:final_prompt}
\small

\begin{quote}
You are a professional annotator of social media posts. Assign a binary code to each post - whether (code 1) or not (code 0) the post reflects basic human values.Values are an expression of the importance of an object, phenomenon, or quality. The Values are listed below:

Self-direction: Importance of independent thought and action, making autonomous decisions based on one’s own judgment, creating and exploring, being curious.

Stimulation: Excitement, novelty, and variety in life, daring. Appreciation of surprises, adventures, trying new things.

Hedonism: Importance of having good time, pleasure and sensuous gratification, 'spoiling' themselves, food, leisure, sex or other personally enjoyable activities.

Achievement: Personal success through demonstrating competence according to social standards, importance of being ambitious, capable, influential, getting social recognition.

Power: Social status and prestige, wealth, control or dominance over people and resources, authority, social power. Importance of being visibly rich and in charge.

Security: Safety, harmony, strong government, stability of relationships, of self, of society and social order, national security, clean, healthy.

Conformity: The restraint of actions, inclinations, and impulses that are likely to upset or harm others and violate social expectations or norms.

Tradition: Respect, commitment, and acceptance of the customs and ideas that traditional culture or religion provides, humble, devout, moderate, subordination.

Benevolence: Preservation and enhancement of the welfare of close people with whom one is in frequent personal contact (e.g., family, friends), helpful, honest, forgiving, responsible toward close others, loyal, true friendship, mature love.

Universalism: Understanding, appreciation, tolerance, and protection for the welfare of all people in the world including strangers; importance of broadmindedness, social justice, equality, world at peace. Importance of nature and environment protection.

When assigning values, prioritize conceptual meaning over surface-level lexical cues.
Don't overestimate emotional texts.
If the value is expressed subtly or implicitly, assign it when there is reasonable textual evidence, rather than requiring explicit keywords.
Do NOT make assumptions about the author and intentions beyond the text.

Power, achievement, hedonism, stimulation, and self-direction primarily focus on personal interests and characteristics. Benevolence, universalism, tradition, conformity, and security are primarily concerned with how one relates socially to others and affects their interests. Power, achievement, tradition, conformity, and security serve to cope with anxiety due to uncertainty in the social and physical world. Hedonism, stimulation, self-direction, universalism, and benevolence express anxiety-free motivations. Security and conformity aim to avoid or overcome actual or potential danger. Self-direction and close values motivate intrinsically rewarding social, intellectual, and emotional opportunities.
\end{quote}
\section{Domain-penalized Jaccard similarity metric $J_{\text{dp}}(E,P)$}
\label{app:jaccard}

Let $E = \{e_1, ..., e_n\}$ denote the set of expert-assigned values and 
$P = \{p_1, ..., p_m\}$ the set of predicted values for a post.

We define pairwise similarity between values as:

\[
s(e_i, p_j) =
\begin{cases}
1, & \text{if } e_i = p_j, \\
0.5, & \text{if } e_i \neq p_j \text{ and both belong to the same higher-order domain}, \\
0, & \text{otherwise}.
\end{cases}
\]

We then compute an optimal bipartite matching between predicted and expert-assigned values:

\[
M(E,P) = \max_{\pi} \sum_{(e_i,p_j)\in \pi} s(e_i,p_j),
\]

where $\pi$ denotes an optimal one-to-one matching between elements of $E$ and $P$.

The final domain-penalized Jaccard similarity score is computed as:

\[
J_{\text{dp}}(E,P)=
\frac{M(E,P)}
{E \cup P},
\]

where $E \cup P$ denotes the number of unique values appearing in the expert-assigned set, the predicted set, or both.

\section{Annotation regimes metrics (Dataset-1, N=1,000)}
\label{app:annotation_regimes}

\subsection{Response of models to prompt refinement} 
\label{app:P-R-F1-across_model-prompts}
\begin{figure}[H]
\centering
\includegraphics[width=\linewidth]{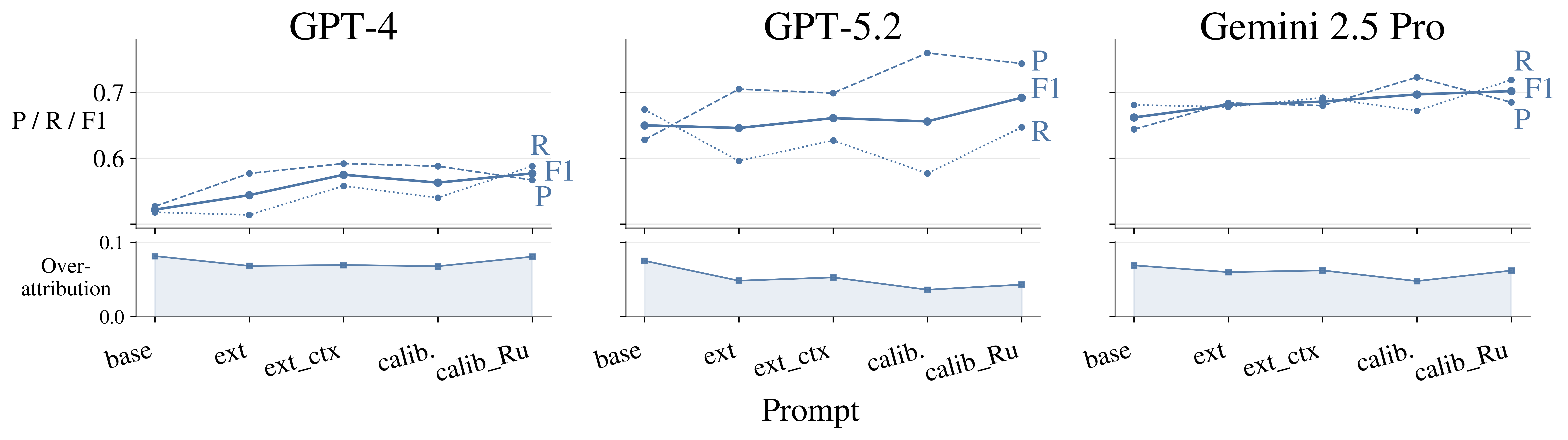}
\caption{Changes in Precision, Recall, F1, and over-attribution rates across stages of prompt refinement.}
\label{fig:P-R-F1-across_model-prompts_blue}
\end{figure}

\subsection{Overall P/R/F1} 
\begin{table}[H]
\centering
\small
\setlength{\tabcolsep}{4pt}

\begin{tabular}{llccc}
\hline
\textbf{Model} & \textbf{Prompt} & \textbf{P} & \textbf{R} & \textbf{F1} \\
\hline

GPT-4 
& Baseline & 0.527 & 0.518 & 0.522 \\
& Extended & 0.577 & 0.514 & 0.544 \\
& Contextualized & 0.592 & 0.558 & 0.575 \\
& Bias-calibrated & 0.588 & 0.540 & 0.563 \\
& Bias-calibrated-RU & 0.567 & 0.588 & 0.577 \\
\hline

GPT-5.2
& Baseline & 0.628 & 0.674 & 0.650 \\
& Extended & 0.705 & 0.596 & 0.646 \\
& Contextualized & 0.699 & 0.627 & 0.661 \\
& Bias-calibrated & 0.760 & 0.577 & 0.656 \\
& Bias-calibrated-RU & 0.744 & 0.647 & 0.692 \\
\hline

Gemini
& Baseline & 0.644 & 0.681 & 0.662 \\
& Extended & 0.684 & 0.678 & 0.681 \\
& Contextualized & 0.680 & 0.692 & 0.686 \\
& Bias-calibrated & 0.723 & 0.672 & 0.697 \\
& Bias-calibrated-RU & 0.685 & 0.719 & 0.702 \\
\hline

\end{tabular}

\caption{Overall Precision, Recall, and F1 across annotation regimes.}
\label{tab:overall_metrics}
\end{table}

\subsection{Per-value metrics} 
\begin{table}[H]
\centering
\small
\setlength{\tabcolsep}{4pt}

\begin{tabular}{lccccccccc}
\hline

& \multicolumn{3}{c}{\textbf{GPT-4}} 
& \multicolumn{3}{c}{\textbf{GPT-5.2}} 
& \multicolumn{3}{c}{\textbf{Gemini}} \\

\cline{2-10}

\textbf{Value}
& \textbf{P} & \textbf{R} & \textbf{F1}
& \textbf{P} & \textbf{R} & \textbf{F1}
& \textbf{P} & \textbf{R} & \textbf{F1} \\

\hline

Self-direction 
& 0.286 & 0.838 & 0.427 
& 0.484 & 0.667 & 0.561 
& 0.459 & 0.703 & 0.555 \\

Stimulation 
& 0.261 & 0.671 & 0.376 
& 0.391 & 0.514 & 0.444 
& 0.450 & 0.643 & 0.529 \\

Hedonism 
& 0.541 & 0.254 & 0.345 
& 0.651 & 0.474 & 0.548 
& 0.694 & 0.565 & 0.623 \\

Achievement 
& 0.664 & 0.654 & 0.659 
& 0.713 & 0.685 & 0.699 
& 0.740 & 0.717 & 0.728 \\

Power 
& 0.271 & 0.471 & 0.344 
& 0.386 & 0.647 & 0.484 
& 0.351 & 0.794 & 0.486 \\

Security 
& 0.780 & 0.337 & 0.471 
& 0.590 & 0.639 & 0.613 
& 0.571 & 0.496 & 0.531 \\

Conformity 
& 0.320 & 0.436 & 0.369 
& 0.431 & 0.400 & 0.415 
& 0.478 & 0.400 & 0.436 \\

Tradition 
& 0.356 & 0.508 & 0.419 
& 0.327 & 0.885 & 0.478 
& 0.377 & 0.852 & 0.523 \\

Benevolence 
& 0.931 & 0.614 & 0.740 
& 0.891 & 0.795 & 0.840 
& 0.926 & 0.787 & 0.851 \\

Universalism 
& 0.375 & 0.434 & 0.402 
& 0.538 & 0.687 & 0.603 
& 0.548 & 0.819 & 0.657 \\

\hline
\end{tabular}

\caption{Per-value Precision, Recall, and F1 scores under Baseline prompt.}
\label{tab:baseline_per_value}
\end{table}

\begin{table}[H]
\centering
\small
\setlength{\tabcolsep}{4pt}

\begin{tabular}{lccccccccc}
\hline

& \multicolumn{3}{c}{\textbf{GPT-4}} 
& \multicolumn{3}{c}{\textbf{GPT-5.2}} 
& \multicolumn{3}{c}{\textbf{Gemini}} \\

\cline{2-10}

\textbf{Value}
& \textbf{P} & \textbf{R} & \textbf{F1}
& \textbf{P} & \textbf{R} & \textbf{F1}
& \textbf{P} & \textbf{R} & \textbf{F1} \\

\hline

Self-direction 
& 0.324 & 0.793 & 0.460 
& 0.625 & 0.495 & 0.553 
& 0.534 & 0.703 & 0.607 \\

Stimulation 
& 0.320 & 0.571 & 0.410 
& 0.500 & 0.371 & 0.426 
& 0.481 & 0.543 & 0.510 \\

Hedonism 
& 0.635 & 0.225 & 0.332 
& 0.772 & 0.292 & 0.424 
& 0.773 & 0.488 & 0.598 \\

Achievement 
& 0.704 & 0.543 & 0.613 
& 0.763 & 0.559 & 0.645 
& 0.724 & 0.701 & 0.712 \\

Power 
& 0.283 & 0.441 & 0.345 
& 0.542 & 0.382 & 0.448 
& 0.429 & 0.706 & 0.533 \\

Security 
& 0.820 & 0.325 & 0.466 
& 0.760 & 0.516 & 0.615 
& 0.708 & 0.500 & 0.586 \\

Conformity 
& 0.339 & 0.364 & 0.351 
& 0.438 & 0.255 & 0.322 
& 0.571 & 0.291 & 0.386 \\

Tradition 
& 0.371 & 0.590 & 0.456 
& 0.521 & 0.820 & 0.637 
& 0.547 & 0.852 & 0.667 \\

Benevolence 
& 0.912 & 0.749 & 0.823 
& 0.906 & 0.804 & 0.852 
& 0.933 & 0.827 & 0.877 \\

Universalism 
& 0.370 & 0.410 & 0.389 
& 0.661 & 0.470 & 0.549 
& 0.573 & 0.807 & 0.670 \\

\hline
\end{tabular}

\caption{Per-value Precision, Recall, and F1 scores under Bias-calibrated prompt.}
\label{tab:calibrated_per_value}
\end{table}

\subsection{Structural alignment Jaccard scores} 
\label{app:jaccard_metrics}
\begin{table}[H]
\centering
\small
\setlength{\tabcolsep}{5pt}
\begin{tabular}{llc}
\hline
\textbf{Model} & \textbf{Prompt} & \textbf{Mean} \\
\hline

GPT-4 
& Baseline & 0.444 \\
& Extended & 0.456 \\
& Contextualized & 0.498 \\
& Bias-calibrated & 0.499 \\
& Bias-calibrated-RU & 0.504 \\
\hline

GPT-5.2
& Baseline & 0.544 \\
& Extended & 0.553 \\
& Contextualized & 0.568 \\
& Bias-calibrated & 0.569 \\
& Bias-calibrated-RU & 0.599 \\
\hline

Gemini
& Baseline & 0.563 \\
& Extended & 0.597 \\
& Contextualized & 0.593 \\
& Bias-calibrated & 0.616 \\
& Bias-calibrated-RU & 0.612 \\
\hline

\end{tabular}

\caption{Domain-penalized Jaccard similarity scores across annotation regimes.}
\label{tab:structural_alignment_mean}
\end{table}

\subsection{Spearman correlations} 
\label{app:correlations_all_regimes}
\begin{table}[H]
\centering
\small
\setlength{\tabcolsep}{6pt}
\begin{tabular}{llc}
\hline
\textbf{Model} & \textbf{Prompt} & \textbf{Spearman $\rho$} \\
\hline

GPT-4 & Baseline & 0.493 \\
& Extended & 0.472 \\
& Contextualized & 0.543 \\
& Bias-calibrated & 0.495 \\
& Bias-calibrated-RU & 0.520 \\
\hline

GPT-5 & Baseline & 0.615 \\
& Extended & 0.609 \\
& Contextualized & 0.624 \\
& Bias-calibrated & 0.621 \\
& Bias-calibrated-RU & 0.638 \\
\hline

Gemini & Baseline & 0.637 \\
& Extended & 0.635 \\
& Contextualized & 0.645 \\
& Bias-calibrated & 0.666 \\
& Bias-calibrated-RU & 0.663 \\
\hline

\end{tabular}

\caption{Spearman correlations between expert and model vote counts across annotation regimes.}
\label{tab:correlations_all_regimes}
\end{table}

\section{Dataset-2, value distribution}
\label{app:expert_votes-dataset2}
\begin{figure}[H]
\centering
\includegraphics[width=0.75\textwidth]{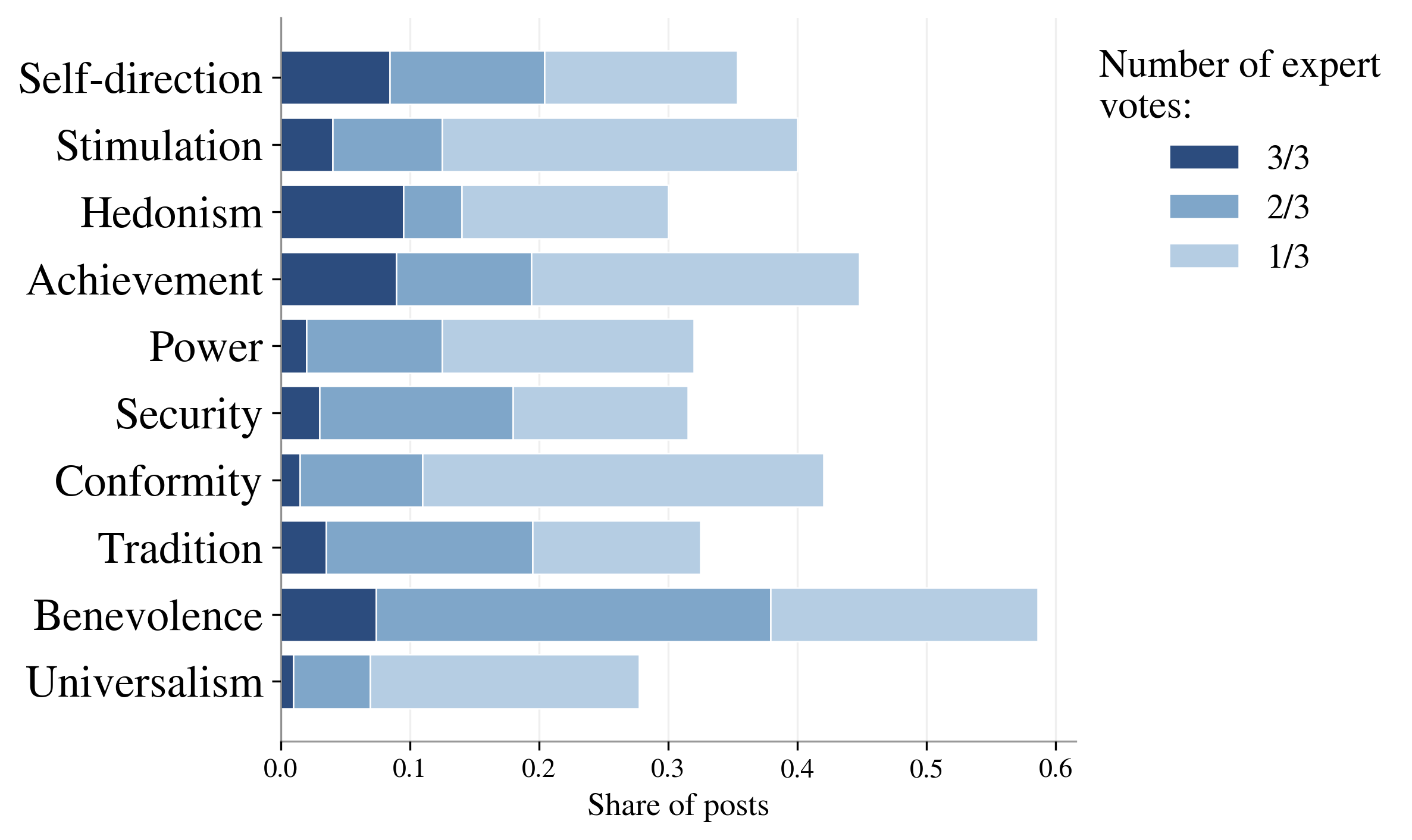}
\caption{Distribution of expert votes across values in Dataset-2. Each value was independently evaluated on a sample of 200 posts.}
\label{fig:expert_votes-dataset2}
\end{figure}
Overall expert agreement in Dataset-2 was low (Fleiss’ k = 0.3), substantially below that observed in Dataset-1 (Fleiss k = 0.6). In addition, the lower proportion of unanimous decisions suggests that Dataset-2 contains a larger number of borderline cases and/or reflects more conservative expert annotation style. This provides a plausible explanation for some of the differences observed between Dataset-1 and Dataset-2, including higher over-attribution rates and lower precision for several values.

\section{Final annotation regime (Gemini bias-calibrated English-language prompt)}

\subsection{Per-value metrics on Dataset-2}
\label{app:dataset2} 
\begin{table}[H]
\centering
\small
\setlength{\tabcolsep}{5pt}

\begin{tabular}{lccccc}
\hline
\textbf{Value} & \textbf{P} & \textbf{R} & \textbf{F1} & \textbf{OA} & \textbf{UA} \\
\hline

Self-direction & 0.49 & 0.90 & 0.64 & 0.24 & 0.17 \\
Stimulation & 0.50 & 0.56 & 0.53 & 0.07 & 0.44 \\
Hedonism & 0.51 & 0.82 & 0.63 & 0.17 & 0.14 \\
Achievement & 0.67 & 0.74 & 0.71 & 0.12 & 0.13 \\
Power & 0.61 & 0.68 & 0.64 & 0.12 & 0.20 \\
Security & 0.45 & 0.94 & 0.61 & 0.33 & 0.06 \\
Conformity & 0.35 & 0.27 & 0.31 & 0.04 & 0.59 \\
Tradition & 0.72 & 0.87 & 0.79 & 0.09 & 0.08 \\
Benevolence & 0.64 & 0.96 & 0.77 & 0.32 & 0.07 \\
Universalism & 0.29 & 0.50 & 0.37 & 0.17 & 0.07 \\

\hline
\end{tabular}

\caption{Precision, Recall, F1, over-attribution rate (OA), and under-attribution rate (UA).}
\label{tab:value_metrics}
\end{table}

\noindent
\textit{Note.} Relative to Dataset-1, Dataset-2 exhibits higher Recall and over-attribution rates, and lower Precision for several values. This pattern is likely influenced by differences in annotation conditions and more conservative expert annotation style, accompanied by low expert agreement.

\subsection{Spearman correlations} \
\label{app:final_regime_ro}
\begin{table}[H]
\centering
\small
\setlength{\tabcolsep}{6pt}

\begin{tabular}{lcc}
\hline
\textbf{Value} & \textbf{Dataset-1} & \textbf{Dataset-2} \\
\hline

Self-direction & 0.634 & 0.635 \\
Stimulation & 0.535 & 0.505 \\
Hedonism & 0.598 & 0.742 \\
Achievement & 0.681 & 0.714 \\
Power & 0.516 & 0.581 \\
Security & 0.597 & 0.681 \\
Conformity & 0.395 & 0.445 \\
Tradition & 0.584 & 0.791 \\
Benevolence & 0.833 & 0.719 \\
Universalism & 0.674 & 0.537 \\

\hline
\end{tabular}

\caption{Spearman correlations between expert and annotation vote counts across values in Dataset-1 and Dataset-2, p < 0.001.}
\label{tab:spearman_datasets}
\end{table}

\section{Error profiles and targeted verification} \

\subsection{Error profiles for over- and under- attributed values on Dataset-2}
\label{app:dataset2_fp_fn}
\begin{figure}[htbp]
\centering
\includegraphics[width=0.6\textwidth]{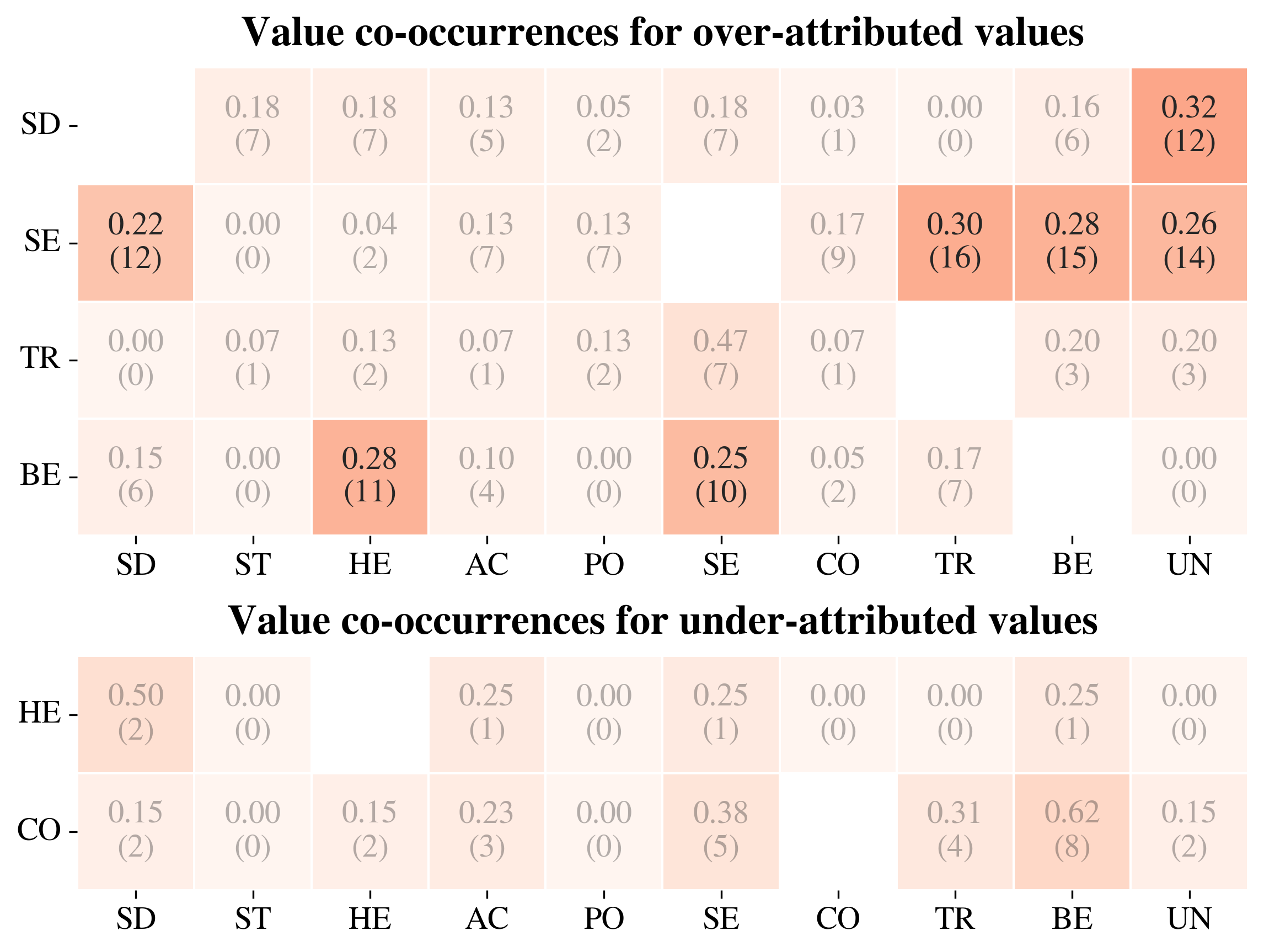}
\caption{Contexts associated with over- and under- attributed values for Dataset-2.}
\label{fig:value_co-occurrences_dataset2}
\end{figure}

\subsection{Targeted expert verification procedure}
\label{app:expert_verification_rules}
\begin{figure}[H]
\centering
\includegraphics[width=0.75\textwidth]{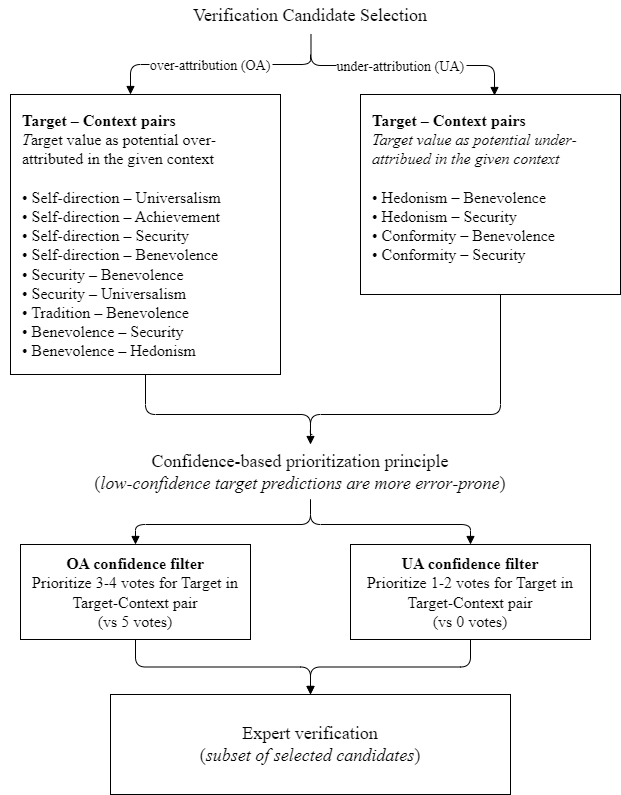}
\caption{Targeted expert verification workflow.}
\label{fig:Figure1_verification_flow}
\end{figure}

\section{Train dataset (N=20,000)} \

\subsection{Gemini annotation results, value distribution}
\label{app:train_data_distribution}
\begin{figure}[H]
\centering
\includegraphics[width=0.75\textwidth]{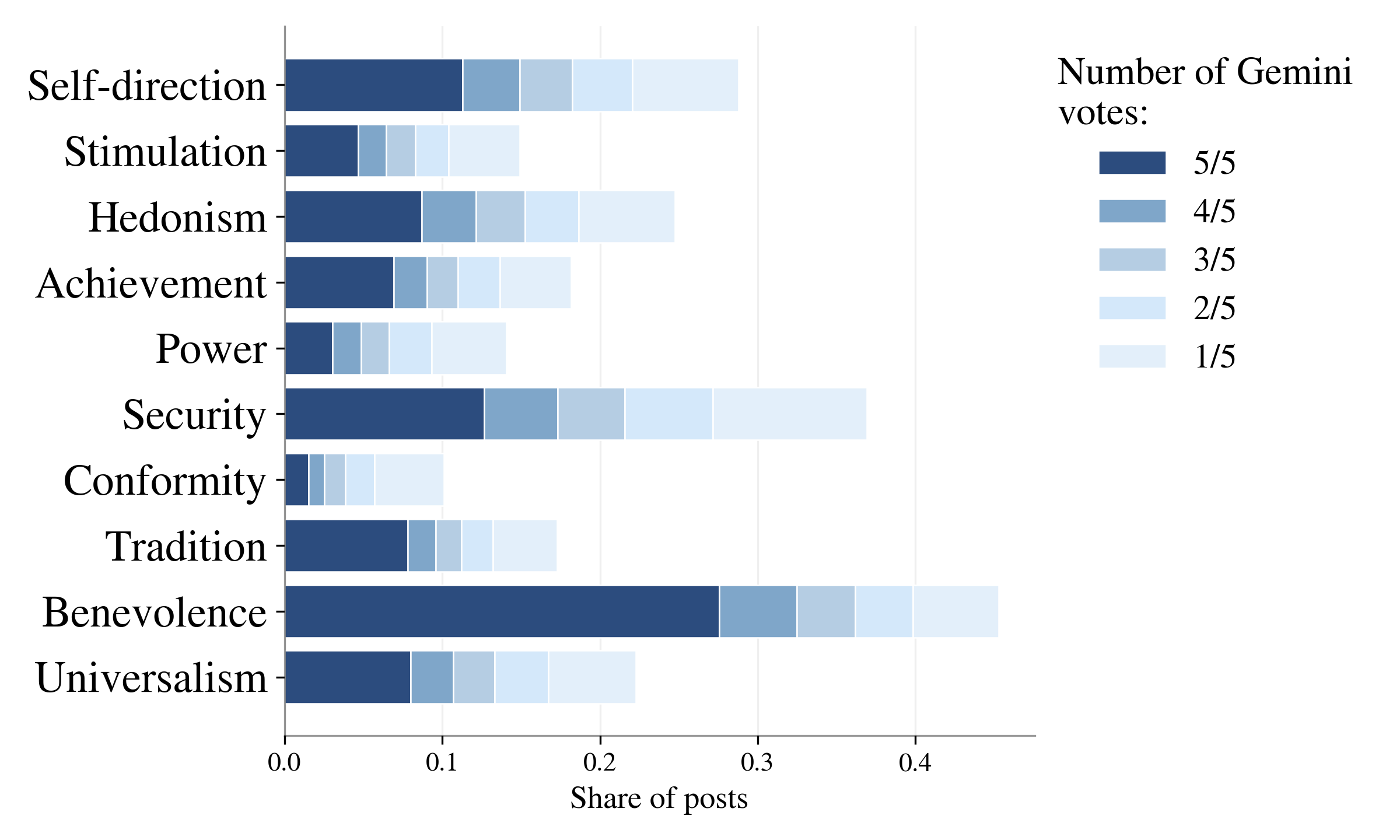}
\caption{Distribution of positive votes across values.}
\label{fig:num_gemini_votes_traindata}
\end{figure}

\subsection{Expert verification and annotation corrections}
\label{app:expert_verification}
\begin{table*}[htbp]
\centering
\small
\setlength{\tabcolsep}{5pt}

\begin{tabular}{llcc}
\hline
\textbf{Target value} & \textbf{Verification rule} & \textbf{Flagged, N} & \textbf{Sampled, N} \\
\hline

Benevolence 
& FP: Benevolence -- Security 
& 358 & 146 \\

Benevolence 
& FP: Benevolence -- Hedonism 
& 291 & 118 \\

Security 
& FP: Security -- Benevolence 
& 690 & 280 \\

Security 
& FP: Security -- Universalism 
& 338 & 136 \\

Self-direction 
& FP: Self-direction -- Achievement 
& 275 & 111 \\

Self-direction 
& FP: Self-direction -- Benevolence 
& 270 & 109 \\

Self-direction 
& FP: Self-direction -- Universalism 
& 223 & 90 \\

Self-direction 
& FP: Self-direction -- Security 
& 181 & 73 \\

Tradition 
& FP: Tradition -- Benevolence 
& 345 & 139 \\

Conformity 
& FN: Conformity -- Benevolence 
& 565 & 229 \\

Conformity 
& FN: Conformity -- Security 
& 536 & 218 \\

Hedonism 
& FN: Hedonism -- Benevolence 
& 921 & 374 \\

Hedonism 
& FN: Hedonism -- Security 
& 232 & 94 \\

\hline
\end{tabular}

\caption{Target--context verification rules and sampled cases for expert review.}
\label{tab:verification_rules}
\end{table*}

\begin{table}[H]
\centering
\small
\setlength{\tabcolsep}{6pt}
\begin{tabular}{lcc}
\hline
\textbf{Target value} & \textbf{Reviewed, N} & \textbf{Corrected share (N)} \\
\hline

Security & 399 & 0.78 (310) \\
Self-direction & 345 & 0.74 (256) \\
Tradition & 138 & 0.59 (81) \\
Benevolence & 257 & 0.45 (116) \\
Hedonism & 439 & 0.26 (113) \\
Conformity & 414 & 0.21 (86) \\

\hline
\end{tabular}
\caption{Expert verification results across target-value groups.}
\label{tab:verification_outcomes}
\end{table}

\noindent
\textit{Note.} Correction shares are reported with respect to the intended direction of verification. For over-attribution rules, a correction is counted only when expert review decreases the target-value label. For under-attribution rules, a correction is counted only when expert review increases the target-value label. Changes in the opposite direction are not counted.

\subsection{Annotation differences between Gemini and GPT-4}
\label{app:train_annotation_differences}
\begin{figure}[H]
\centering
\includegraphics[width=\linewidth]{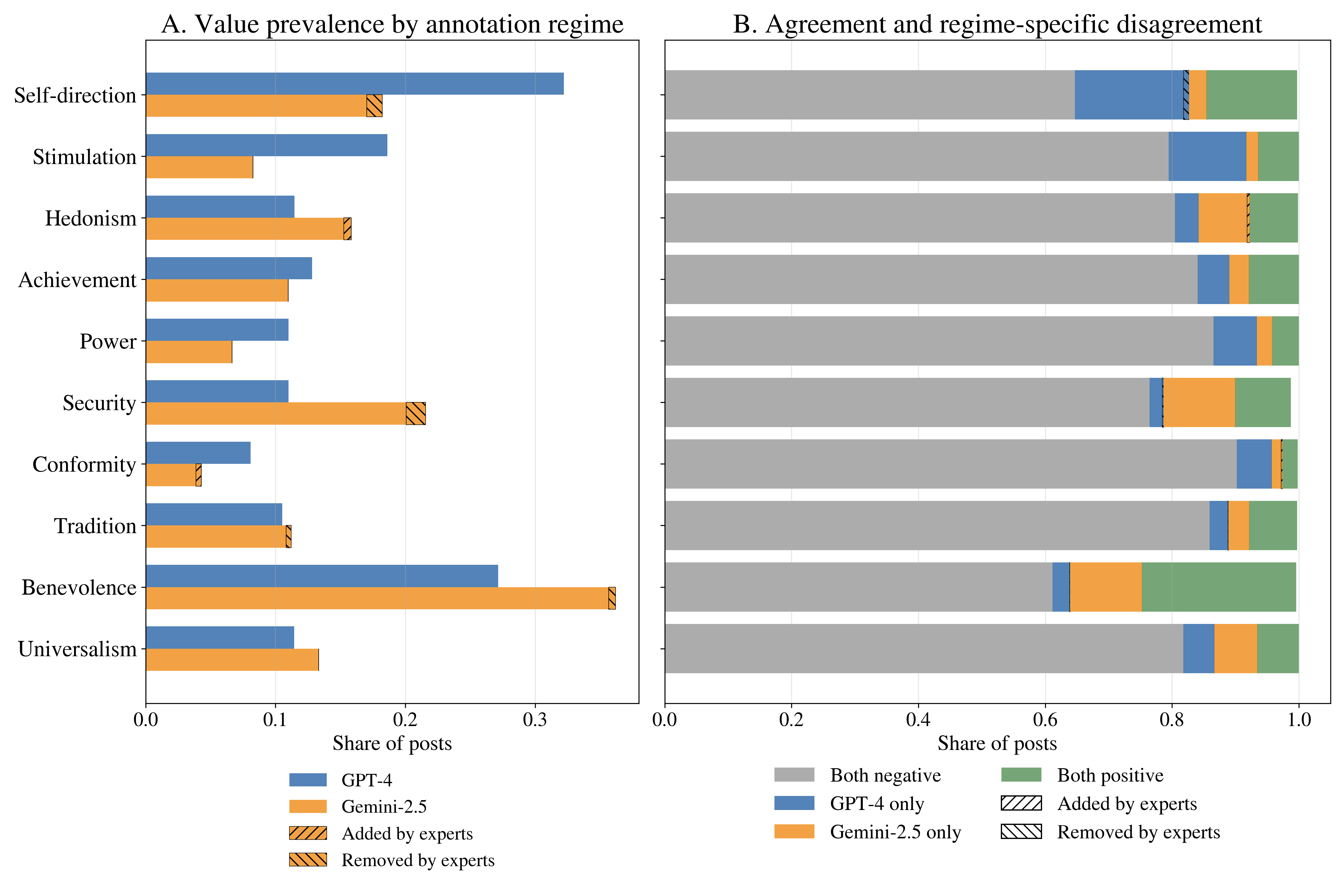}
\caption{Annotation differences on train dataset: Gemini 2.5 (bias-calibrated prompt) vs GPT-4 (baseline prompt) results.}
\label{fig:20000_annotation_differences}
\end{figure}
\noindent
\textit{Note.} The figure illustrates how alternative annotation regimes operationalize value categories differently. Panel A shows value prevalence differences between the earlier GPT-4 annotation regime and the current calibrated Gemini regime. Panel B shows agreement structure and regime-specific disagreements. Expert corrections were applied only to the Gemini regime. GPT-4 results are shown for historical comparison

\section{Encoder-transfer results}

\subsection{Encoder training details}
\label{app:xlm_training}
We fine-tuned the last two layers of XLM-RoBERTa-large using mean pooling, linear multi-label classification head, and binary cross-entropy (BCE) loss. Other parameters were: batch size 16, maximum sequence length 512, AdamW optimizer with a learning rate of 2e-5, and early stopping based on validation loss and PR-AUC (6 epochs). 
Training was performed on a server equipped with an NVIDIA GeForce RTX 2080 Ti GPU, 48 CPU cores, and 128 GB of RAM.

\subsection{XLM-RoBERTa performance against Gemini annotations, on test dataset (N=4,000)}
\label{app:xlm_results}
\begin{table}[H]
\centering
\small
\setlength{\tabcolsep}{6pt}
\begin{tabular}{lc}
\hline
\textbf{Value} & \textbf{AP} \\
\hline

Self-direction & 0.752 \\
Stimulation & 0.633 \\
Hedonism & 0.726 \\
Achievement & 0.794 \\
Power & 0.473 \\
Security & 0.686 \\
Conformity & 0.429 \\
Tradition & 0.755 \\
Benevolence & 0.930 \\
Universalism & 0.702 \\
\hline
\end{tabular}
\caption{Precision--Recall AUC (Average Precision, AP).}
\label{tab:ap_values}
\end{table}

\begin{figure}[H]
\centering
\includegraphics[width=0.5\textwidth]{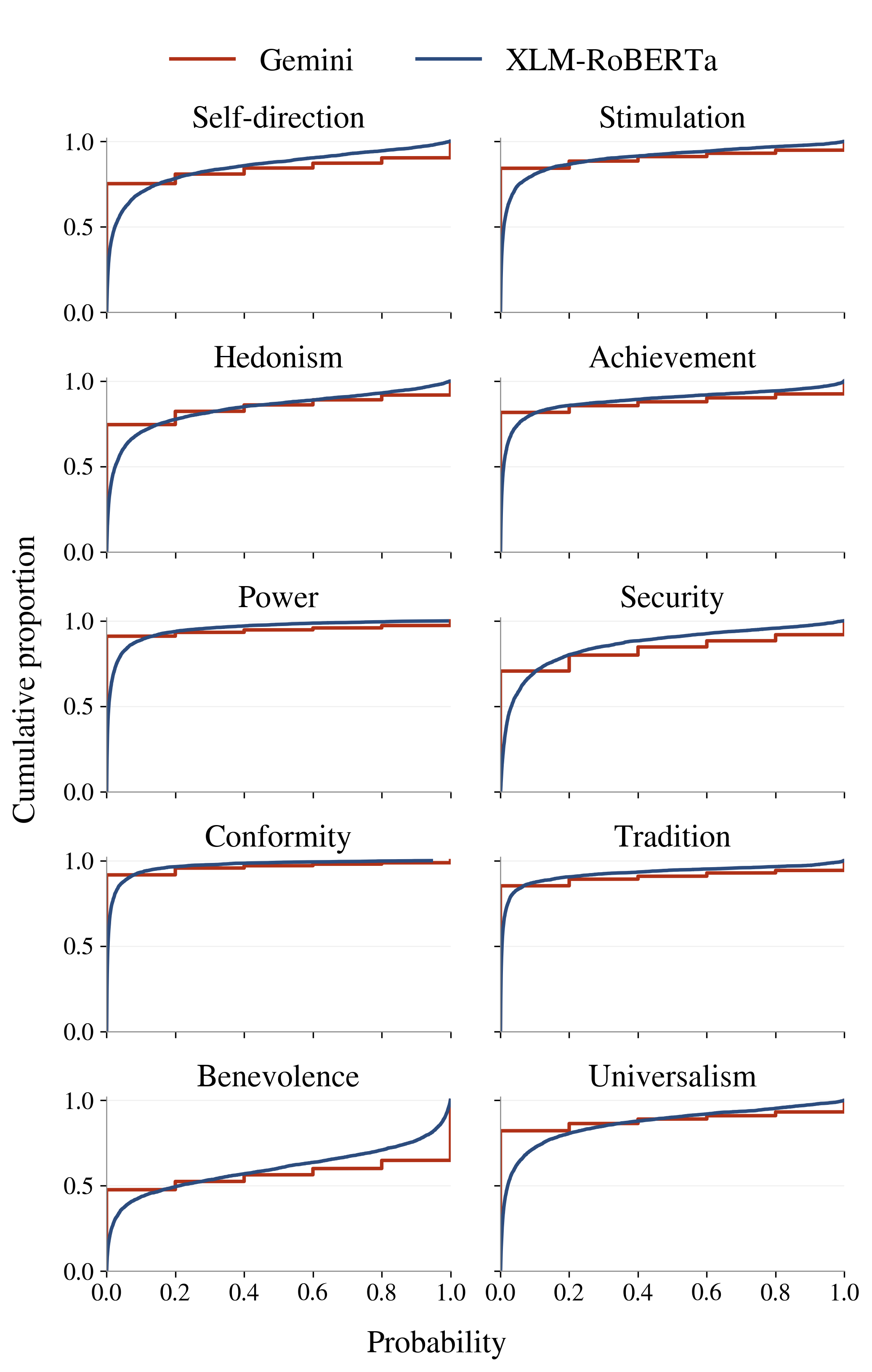}
\caption{Cumulative distribution functions.}
\label{fig:bert-llm-testdata-cumulative}
\end{figure}

\section{Sensitivity analysis}
\label{app:sensitivity}
\begin{table*}[htbp]
\centering
\small
\begin{tabular}{lrrrrrr}
\toprule
Value & GPT mean & Gemini mean & Diff. & 95\% CI & Rel. diff. & Share Gemini $>$ GPT \\
\midrule
Self-direction & 0.351 & 0.209 & -0.142 & [-0.146, -0.137] & -40.5\% & 0.000 \\
Stimulation    & 0.173 & 0.104 & -0.070 & [-0.081, -0.058] & -40.2\% & 0.000 \\
Hedonism       & 0.143 & 0.189 &  0.047 & [ 0.028,  0.066] &  32.7\% & 0.727 \\
Achievement    & 0.128 & 0.124 & -0.003 & [-0.008,  0.001] &  -2.7\% & 0.440 \\
Power          & 0.085 & 0.052 & -0.033 & [-0.041, -0.025] & -38.6\% & 0.139 \\
Security       & 0.116 & 0.191 &  0.075 & [ 0.072,  0.078] &  64.9\% & 1.000 \\
Conformity     & 0.061 & 0.045 & -0.016 & [-0.017, -0.015] & -26.3\% & 0.014 \\
Tradition      & 0.090 & 0.127 &  0.037 & [ 0.033,  0.041] &  41.5\% & 1.000 \\
Benevolence    & 0.314 & 0.423 &  0.109 & [ 0.097,  0.121] &  34.7\% & 1.000 \\
Universalism   & 0.144 & 0.137 & -0.007 & [-0.010, -0.004] &  -4.8\% & 0.311 \\
\bottomrule
\end{tabular}
\caption{Differences in monthly aggregated predicted levels (raw probability) of value expression between GPT-based and Gemini-based predictions, moving-block bootstrap 95\% CIs.}
\label{app:sensitivity_level_differences}
\end{table*}

\begin{table*}[htbp]
\centering
\small
\begin{tabular}{lccccc}
\toprule
Value & Directional agreement & $\beta$ & 95\% HAC CI & RMSE$_\Delta$ & $R^2$ \\
\midrule
Self-direction & 88.5\% & 0.72 & [0.66, 0.78] & 0.027 & 0.84 \\
Stimulation    & 85.6\% & 0.72 & [0.64, 0.79] & 0.029 & 0.77 \\
Hedonism       & 78.8\% & 1.02 & [0.84, 1.21] & 0.060 & 0.45 \\
Achievement    & 86.5\% & 0.82 & [0.71, 0.93] & 0.029 & 0.78 \\
Power          & 88.0\% & 0.76 & [0.68, 0.83] & 0.032 & 0.87 \\
Security       & 87.0\% & 0.90 & [0.83, 0.97] & 0.026 & 0.93 \\
Conformity     & 71.2\% & 0.73 & [0.49, 0.98] & 0.042 & 0.34 \\
Tradition      & 85.6\% & 0.90 & [0.85, 0.96] & 0.034 & 0.89 \\
Benevolence    & 84.6\% & 0.88 & [0.80, 0.96] & 0.028 & 0.78 \\
Universalism   & 70.7\% & 0.50 & [0.25, 0.74] & 0.060 & 0.26 \\
\bottomrule
\end{tabular}
\caption{Agreement in monthly temporal dynamics between GPT-based and Gemini-based predictions. Directional agreement is the percentage of months in which first differences of value-specific z-standardized monthly scores have the same sign. $\beta$ is the coefficient from per-value monthly regressions of $\Delta z_{\mathrm{Gemini}}$ on $\Delta z_{\mathrm{GPT}}$. CIs use HAC standard errors.}

\end{table*}
\clearpage
\section{Implementation details}
\label{app:implementation}
All analyses were implemented in Python 3.10.14. Data processing and aggregation used pandas 2.2.3 and NumPy 2.1.3. Model evaluation and sensitivity analysis used scikit-learn 1.6.1, SciPy 1.15.0, and statsmodels 0.14.6. Encoder training used PyTorch 2.5.1 with CUDA 12.4 and the Hugging Face Transformers library 4.48.0.

\end{document}